\documentclass[10pt,twocolumn,letterpaper]{article}

\usepackage[pagenumbers]{cvpr}

\usepackage{graphicx}
\usepackage{amsmath}
\usepackage{amssymb}
\usepackage{booktabs}

\usepackage{microtype}
\usepackage{pifont}
\usepackage{enumitem}

\usepackage[pagebackref=false,breaklinks=true,letterpaper=true,colorlinks,bookmarks=false,citecolor=cyan]{hyperref}

\setitemize{noitemsep,topsep=0pt,parsep=0pt,partopsep=0pt}

\newcommand{\cmark}{\ding{51}}
\newcommand{\xmark}{\ding{55}}

\def\VspaceL{\vspace{-0.40cm}}
\def\VspaceS{\vspace{-0.30cm}}

\def\cvprPaperID{1083}
\def\confName{CVPR}
\def\confYear{2022}

\begin{document}

\title{MHFormer: Multi-Hypothesis Transformer for 3D Human Pose Estimation}

\author{
  Wenhao Li\textsuperscript{1} \quad
  Hong Liu\textsuperscript{1,}\thanks{Corresponding author: hongliu@pku.edu.cn. 
  This work is supported by National Key R\&D Program of China (No. 2020AAA0108904), Science and Technology Plan of Shenzhen (No. JCYJ20200109140410340).} \quad 
  Hao Tang\textsuperscript{2} \quad 
  Pichao Wang\textsuperscript{3} \quad 
  Luc Van Gool\textsuperscript{2} \\
  \textsuperscript{1}Key Laboratory of Machine Perception, Shenzhen Graduate School, Peking University \\
  \textsuperscript{2}Computer Vision Lab, ETH Zurich \quad
  \textsuperscript{3}Alibaba Group \\
  {\tt\small \{wenhaoli,hongliu\}@pku.edu.cn} \quad \\ {\tt\small\{hao.tang,vangool\}@vision.ee.ethz.ch} \quad 
  {\tt\small pichao.wang@alibaba-inc.com}
}

\maketitle
  
\begin{abstract}
Estimating 3D human poses from monocular videos is a challenging task due to depth ambiguity and self-occlusion. Most existing works attempt to solve both issues by exploiting spatial and temporal relationships. However, those works ignore the fact that it is an inverse problem where multiple feasible solutions (i.e., hypotheses) exist. To relieve this limitation, we propose a Multi-Hypothesis Transformer (MHFormer) that learns spatio-temporal representations of multiple plausible pose hypotheses. In order to effectively model multi-hypothesis dependencies and build strong relationships across hypothesis features, the task is decomposed into three stages: (i) Generate multiple initial hypothesis representations; (ii) Model self-hypothesis communication, merge multiple hypotheses into a single converged representation and then partition it into several diverged hypotheses; (iii) Learn cross-hypothesis communication and aggregate the multi-hypothesis features to synthesize the final 3D pose. Through the above processes, the final representation is enhanced and the synthesized pose is much more accurate. Extensive experiments show that MHFormer achieves state-of-the-art results on two challenging datasets: Human3.6M and MPI-INF-3DHP. Without bells and whistles, its performance surpasses the previous best result by a large margin of 3\% on Human3.6M. Code and models are available at \url{https://github.com/Vegetebird/MHFormer}. 
\end{abstract}
  
\section{Introduction}
3D human pose estimation (HPE) from monocular videos is a fundamental vision task with a wide range of applications, such as action recognition~\cite{liu2017enhanced,liu2018recognizing,wang2018depth}, human-computer interaction~\cite{errity2016human}, and augmented/virtual reality~\cite{mehta2017vnect}. 
This task is typically solved by dividing it into two decoupled subtasks, \textit{i.e.}, 2D pose detection to localize the keypoints on the image plane, followed by 2D-to-3D lifting to infer joint locations in the 3D space from 2D keypoints. 
Despite their impressive performance~\cite{martinez2017simple,pavllo20193d,chen2021anatomy,gong2021poseaug}, 
it remains an inherently ill-posed problem because of self-occlusion and depth ambiguity in 2D representations. 

\begin{figure}[tb]
  \centering
  \includegraphics[width=1.00\linewidth]{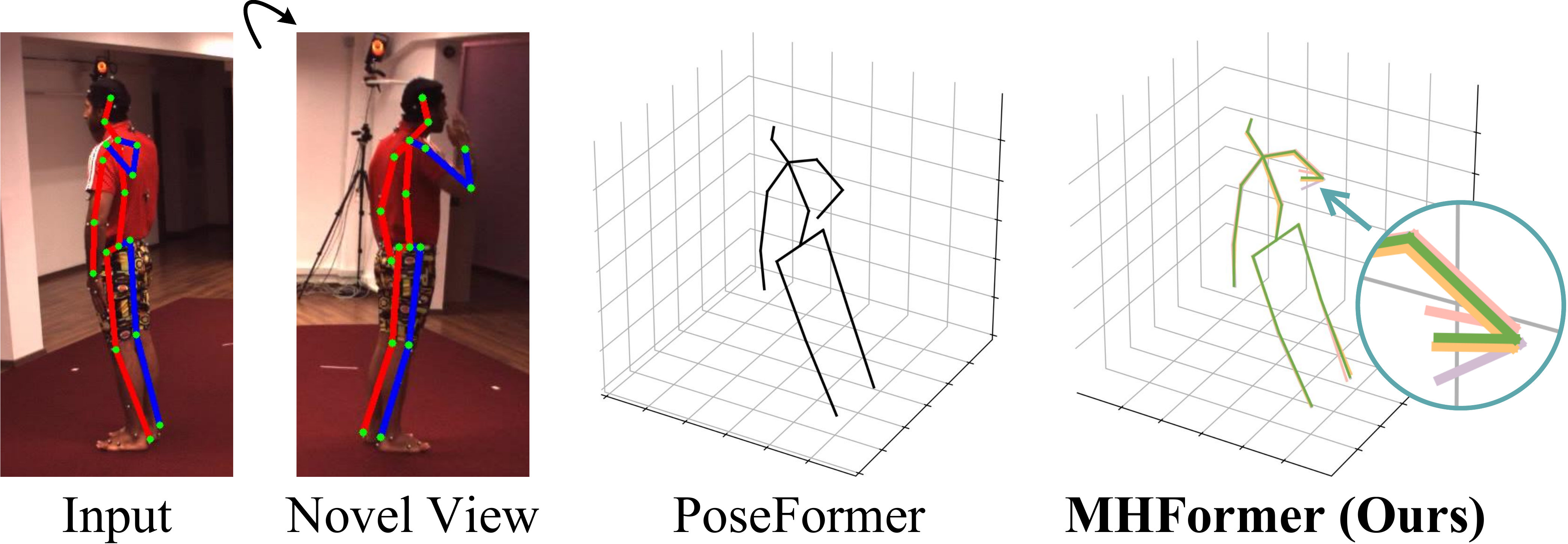}
  \caption
  {Given a frame with occluded body parts (right arm and elbow), a recent state-of-the-art 3D HPE method, PoseFormer~\cite{poseformer}, outputs a single solution that is inconsistent with the 2D input. 
  In contrast, our MHFormer generates multiple plausible hypotheses (different colors) consistent with the 2D evidence and finally synthesizes a more accurate 3D pose (green). 
  For easy comparison, the input frame is shown at a novel viewpoint.}
  \label{fig:motivation}
  \VspaceL
\end{figure}

To alleviate such issues, most  methods~\cite{cai2019exploiting,wang2020motion,hu2021conditional,poseformer} focus on exploring spatial and temporal relationships. 
They either employ graph convolutional networks to estimate 3D poses with a spatio-temporal graph representation of human skeletons~\cite{cai2019exploiting,wang2020motion,hu2021conditional} or apply a pure Transformer-based model to capture spatial and temporal information from 2D pose sequences~\cite{poseformer}. 
Yet, the 2D-to-3D lifting from monocular videos is an inverse problem~\cite{bishop1994mixture} where multiple feasible solutions (\emph{i.e.}, hypotheses) exist due to its ill-posed nature given the missing depth~\cite{li2020weakly}. 
Those approaches ignore this problem and only estimate a single solution, which often leads to unsatisfactory results, especially when the person is severely occluded (see Figure~\ref{fig:motivation}). 

\begin{figure*}[htb]
  \centering
  \includegraphics[width=0.98\linewidth]{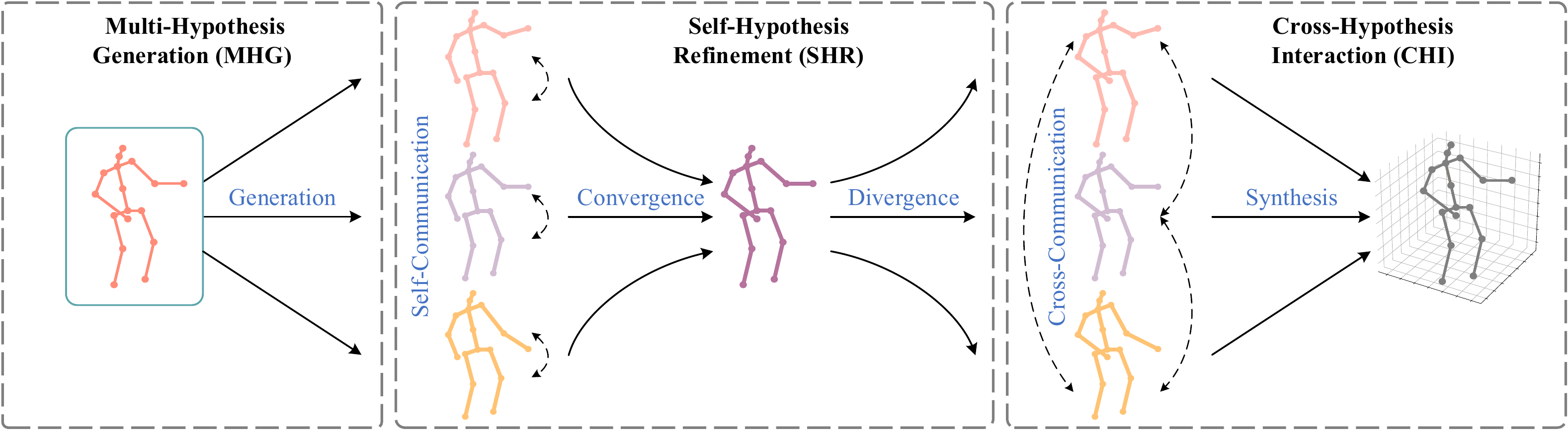}
  \caption
  {The proposed MHFormer constructs a three-stage framework by starting from generating multiple initial representations and then communicating among them in both independent and mutual ways to synthesize a more precise estimation.
  For easy illustration, we only show the process of a single-frame 2D pose as input.}
  \label{fig:pipline}
  \VspaceL
\end{figure*}

Recently, a couple of methods~\cite{khirodkar2021multi,sharma2019monocular,li2019generating,wehrbein2021probabilistic} that generate multiple hypotheses have been proposed for the inverse problem. 
They often rely on the one-to-many mapping by adding multiple output heads to an existing architecture with a shared feature extractor, while failing to build the relationships among the features of different hypotheses. 
That is an important shortcoming, as such ability is vital to improve the expressiveness and the performance of the model. 
In view of the ambiguous inverse problem of 3D HPE, we argue that it is more reasonable to conduct a one-to-many mapping first and then a many-to-one mapping with various intermediate hypotheses, as this way can enrich the diversity of features and produce a better synthesis for the final 3D pose. 

To this end, we present Multi-Hypothesis Transformer (MHFormer), a novel Transformer-based method for 3D HPE from monocular videos. 
The key insight is to allow the model to learn spatio-temporal representations of diverse pose hypotheses. 
To accomplish this, we introduce a three-stage framework that starts from generating multiple initial representations and gradually communicates across them to synthesize a more accurate prediction, as shown in Figure~\ref{fig:pipline}. 
This framework more effectively models multi-hypothesis dependencies while also building stronger relationships among hypothesis features. 
Specifically, in the first stage, a Multi-Hypothesis Generation (MHG) module is built to model the intrinsic structure information of human joints and generate several multi-level features in the spatial domain. 
Those features contain diverse semantic information in different depths from shallow to deep and hence can be regarded as initial representations of multiple hypotheses. 

Next, we propose two novel modules to model temporal consistencies and enhance those coarse representations in the temporal domain, which have not been explored by the existing works that generate multiple hypotheses.
In the second stage, a Self-Hypothesis Refinement (SHR) module is proposed to refine every single-hypothesis feature. 
The SHR consists of two new blocks.
The first block is a multi-hypothesis self-attention (MH-SA) which models single-hypothesis dependencies independently to construct self-hypothesis communication, enabling message passing within each hypothesis for feature enhancement. 
The second block is a hypothesis-mixing multi-layer perceptron (MLP) that exchanges information across hypotheses. 
The multiple hypotheses are merged into a single converged representation, and then this representation is partitioned into several diverged hypotheses. 

Although those hypotheses are refined by SHR, the connections across different hypotheses are not strong enough since the MH-SA in the SHR only passes intra-hypothesis information. 
To address this issue, in the last stage, a Cross-Hypothesis Interaction (CHI) module models interactions among multi-hypothesis features. 
Its key component is the multi-hypothesis cross-attention (MH-CA), which captures mutual multi-hypothesis correlations to build cross-hypothesis communication, enabling message passing among hypotheses for better interaction modeling. 
Subsequently, a hypothesis-mixing MLP is used to aggregate the multiple hypotheses to synthesize the final prediction. 

With the proposed MHFormer, multi-hypothesis spatio-temporal feature hierarchies are explicitly incorporated into Transformer models, where the multiple hypothesis information of body joints can be independently and mutually processed in an end-to-end manner. 
As a result, the representation ability is potentially enhanced and the synthesized pose is much more accurate. 
Our contributions are summarized as follows:
\begin{itemize}
  \item We present a new Transformer-based method, called Multi-Hypothesis Transformer (MHFormer), for 3D HPE from monocular videos. 
  MHFormer can effectively learn spatio-temporal representations of multiple pose hypotheses in an end-to-end manner. 
  \item We propose to communicate among multi-hypothesis features both independently and mutually, providing powerful self-hypothesis and cross-hypothesis message passing, and strong relationships among hypotheses. 
  \item Our MHFormer achieves state-of-the-art performance on two challenging datasets for 3D HPE, significantly outperforming PoseFormer~\cite{poseformer} by 3\% with 1.3 $mm$ error reduction on Human3.6M~\cite{ionescu2013human3}. 
\end{itemize}

\begin{figure*}[htb]
  \centering
  \includegraphics[width=1.00\linewidth]{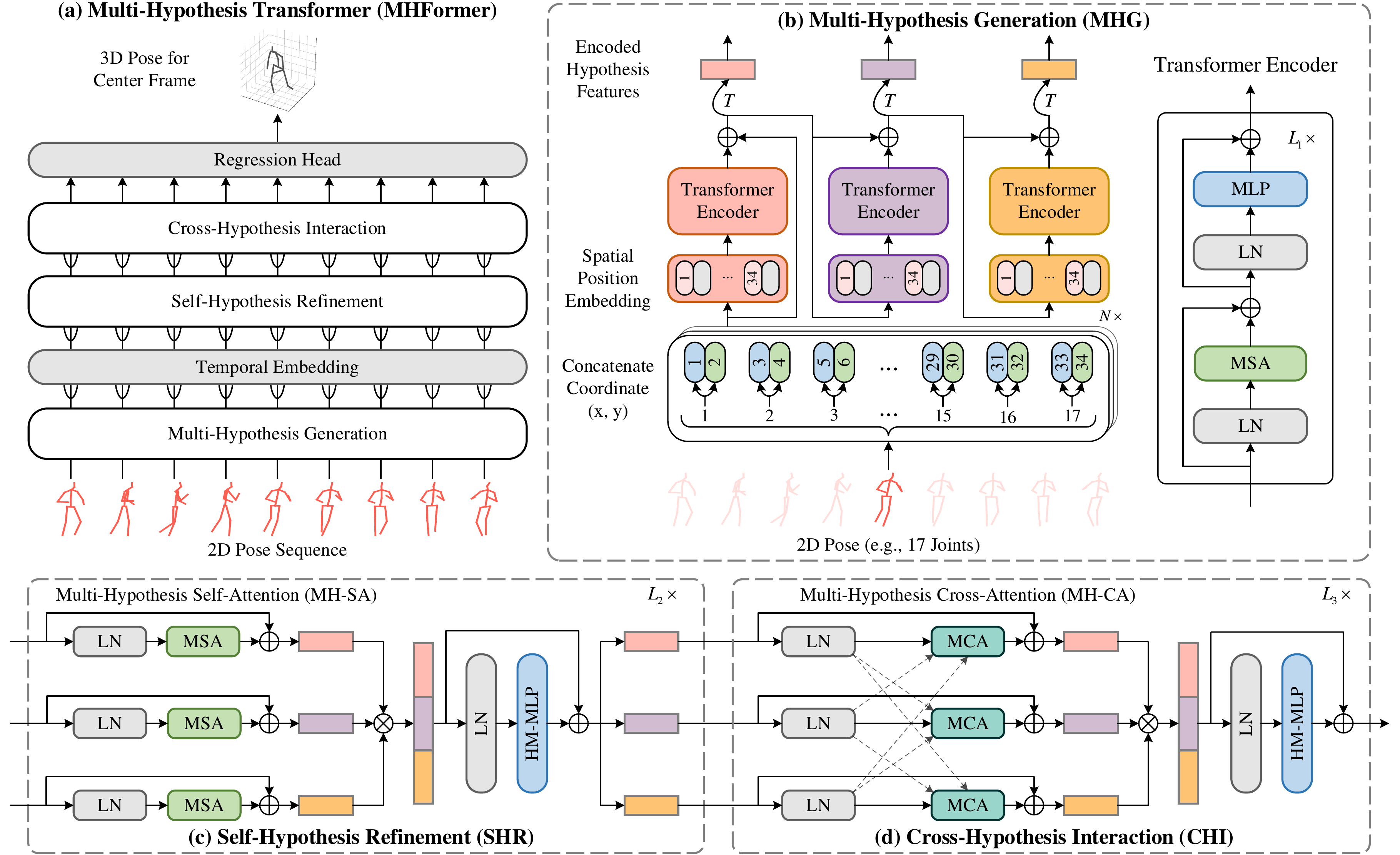}
  \caption
  {\textbf{(a)} Overview of the proposed Multi-Hypothesis Transformer (MHFormer). 
  \textbf{(b)} Multi-Hypothesis Generation (MHG) module extracts the intrinsic structure information of human joints within each frame and generates multiple hypothesis representations. 
  $N$ is the number of input frames and $T$ is the matrix transposition. 
  \textbf{(c)} Self-Hypothesis Refinement (SHR) module is used to refine single-hypothesis features. 
  \textbf{(d)} Cross-Hypothesis Interaction (CHI) module following SHR enables interactions among multi-hypothesis features. 
  }
  \label{fig:overview}
  \VspaceL
\end{figure*}

\section{Related Work} 
\noindent \textbf{3D Human Pose Estimation.}
Existing single-view 3D pose estimation methods can be divided into two mainstream types: one-stage approaches and two-stage ones. 
One-stage approaches directly infer 3D poses from input images without intermediate 2D pose representations~\cite{li20143d,pavlakos2017coarse,sun2018integral,ma2021context}, while two-stage ones first obtain 2D keypoints from pretrained 2D pose detections and then feed them into a 2D-to-3D lifting network to estimate 3D poses. 
Benefiting from the excellent performance of 2D human pose estimation, this 2D-to-3D pose lifting method can efficiently and accurately regress 3D poses using detected 2D keypoints. 
For instance, SimpleBaseline~\cite{martinez2017simple} proposes a fully-connected residual network to lift 2D keypoints to 3D joint locations from a single frame. 
Anatomy3D~\cite{chen2021anatomy} decomposes the task into bone direction and bone length predictions to ensure temporal consistency over a sequence. 
Despite the promising results achieved by using temporal correlations from fully convolutional~\cite{pavllo20193d,liu2020attention,chen2021anatomy}
or graph-based~\cite{cai2019exploiting,wang2020motion,hu2021conditional} architectures, these methods are less efficient in capturing global-context information across frames. 

\noindent \textbf{Vision Transformers.}
Recently, Transformer~\cite{Attention} equipped with the powerfully global self-attention mechanism has received increasingly research interest in the computer vision community~\cite{he2021transreid,yang2021transformer,liu2021swin,lin2021end,han2021transformer}. 
For the basic image classification
task, ViT~\cite{dosovitskiy2020image} is proposed to apply a standard Transformer architecture directly to sequential image patches. 
For the pose estimation task, PoseFormer~\cite{poseformer} applies a pure Transformer to capture human joint correlations and temporal dependencies. 
Strided Transformer~\cite{strided} introduces a Transformer-based architecture with strided convolutions to lift a long 2D pose sequence to a single 3D pose. 
Our work is inspired by them and similarly uses the Transformer as the basic architecture. 
But we do not just utilize a simple architecture with a single representation; instead, the seminal ideas of multi-hypothesis and multi-level feature hierarchies are connected within Transformers, which makes the model not only expressive but also strong. 
Besides, a cross-attention mechanism is introduced for effective multi-hypothesis learning. 

\noindent \textbf{Multi-Hypothesis Methods.}
Single-view 3D HPE is ill-posed and therefore assuming only a single solution might be sub-optimal. 
Several works generate diverse hypotheses for the inverse problem and achieve substantial performance gains~\cite{jahangiri2017generating,li2020weakly,wehrbein2021probabilistic,oikarinen2021graphmdn}. 
For example,
Jahangiri \emph{et al.}~\cite{jahangiri2017generating} generated multiple 3D pose candidates consistent with 2D keypoints via a compositional model and anatomical constraints. 
Wehrbein \emph{et al.}~\cite{wehrbein2021probabilistic} modeled the posterior distribution of 3D pose hypotheses with normalized flows. 
Unlike these works that focus on a one-to-many mapping, we learn a one-to-many mapping first and then a many-to-one mapping, which allows for the effective modeling of different features corresponding to the various hypotheses to improve the representation ability. 

\section{Multi-Hypothesis Transformer}
The overview of the proposed MHFormer is depicted in Figure~\ref{fig:overview} (a). 
Given a consecutive 2D pose sequence estimated by an off-the-shelf 2D pose detector from a video, our method aims to reconstruct the 3D pose of the center frame by making full use of spatial and temporal information in the multi-hypothesis feature hierarchies. 
To achieve our proposed three-stage framework, MHFormer is built upon (i) three major modules: Multi-Hypothesis Generation (MHG), Self-Hypothesis Refinement (SHR), and Cross-Hypothesis Interaction (CHI), and (ii) two auxiliary modules: temporal embedding and regression head. 

\subsection{Preliminary}
In this work, we adopt a Transformer-based architecture since it performs well in long-range dependency modeling. 
We first give a brief description of the basic components in the Transformer~\cite{Attention}, including a multi-head self-attention (MSA) and a multi-layer perceptron (MLP). 

\noindent \textbf{MSA.}
In the MSA, the inputs $x {\in} \mathbb{R}^{n \times d}$ are linearly mapped to queries $Q {\in} \mathbb{R}^{n \times d}$, keys $K {\in} \mathbb{R}^{n \times d}$, and values $V {\in} \mathbb{R}^{n \times d}$, where $n$ is the sequence length, and $d$ is the dimension. 
The scaled dot-product attention can be computed by:
\begin{equation}
  \operatorname{Attention}(Q, K, V)=\operatorname{Softmax}\left({Q K^{T}}/{\sqrt{d}}\right) V.
\end{equation}
MSA splits the queries, keys, and values for $h$ times as well as performs the attention in parallel.  
Then, the outputs of $h$ attention heads are concatenated. 

\noindent \textbf{MLP.}
The MLP consists of two linear layers, which are used for non-linearity and feature transformation: 
\begin{equation}
  \label{equ:mlp}
  \operatorname{MLP}(x)=\sigma\left(x W_{1}+b_{1}\right) W_{2}+b_{2},
\end{equation}
where $\sigma$ denotes the GELU activation function, $W_{1} {\in} \mathbb{R}^{d \times d_{m}}$ and $W_{2} {\in} \mathbb{R}^{d_{m} \times d}$ are the weights of the two linear layers respectively, and $b_{1} {\in} \mathbb{R}^{d_{m}}$ and $b_{2} {\in} \mathbb{R}^{d}$ are the bias terms. 

\subsection{Multi-Hypothesis Generation}
In the spatial domain, we address the inverse problem by explicitly designing a cascaded Transformer-based architecture to generate multiple features in different depths of the latent space. 
To this end, MHG is introduced to model the human joint relations and initialize the multi-hypothesis representations (see Figure~\ref{fig:overview} (b)). 
Suppose there are $M$ different hypotheses and $L_{1}$ layers in the MHG, it takes a sequence of 2D poses $X {\in} \mathbb{R}^{N  \times J \times 2}$ with $N$ video frames and $J$ body joints as input and outputs multiple hypotheses $[{X}^{1}_{L_{1}}, {X}^{2}_{L_{1}}, ..., {X}^{M}_{L_{1}}]$, where ${X}^{m}_{L_{1}} {\in} \mathbb{R}^{(J \cdot 2) \times N}$ is the $m$-th hypothesis.

More specifically, we concatenate the $(x, y)$ coordinates of joints for each frame to $\bar{X} {\in} \mathbb{R}^{(J \cdot 2) \times N}$, retain their spatial information of joints via a learnable spatial position embedding $E^{m}_{SPos} {\in} \mathbb{R}^{(J \cdot 2) \times N}$, and feed the embedded features into the encoders of the MHG. 
To encourage gradient propagation, a skip residual connection is applied between the original input and output features from the encoder. 
These procedures can be formulated as: 
\begin{align}
  \begin{split}
  X^{m}_{0} &= \operatorname{LN}(X^{m}) + E^{m}_{SPos}, \\
  X_{l}^{\prime m} &= X^{m}_{l-1}+\operatorname{MSA}^{m}(\operatorname{LN}(X^{m}_{l-1}), 
  \\
  {X}^{\prime \prime m}_{l} &= X_{l}^{\prime m} + \operatorname{MLP}^{m}(\operatorname{LN}(X_{l}^{\prime m}), \\
  {X}^{m}_{L_{1}} &= X^{m} + \operatorname{LN}({X}^{\prime \prime m}_{L_{1}}), 
  \end{split}
\end{align}
where $\operatorname{LN}(\cdot)$ is the LayerNorm layer, $l {\in} [1, ..., L_{1}]$ is the index of MHG layers, $X^{1}{=}\bar{X}$, and $X^{m}{=}X^{m-1}_{L_{1}}$ ($m {>} 1$). 
The outputs of the MHG (\textit{i.e.}, ${X}^{m}_{L_{1}}$) are multi-level features containing diverse semantic information. 
Therefore, those features can be regarded as initial representations of different pose hypotheses and need to be further enhanced. 

\subsection{Temporal Embedding}
The MHG helps to generate initial multi-hypothesis features in the spatial domain, whereas the capabilities of such features are not strong enough. 
Considering this limitation, we propose to build relationships across hypothesis features and capture temporal dependencies in the temporal domain with two carefully designed modules: an SHR module followed by a CHI module (see Figure~\ref{fig:overview} (c) and (d)). 

In order to exploit temporal information, we should first convert the spatial domain into the temporal domain. 
For this purpose, the encoded hypothesis features ${X}^{m}_{L_{1}}$ of each frame are embedded to the high-dimensional features $\widetilde{Z}^{m} {\in} \mathbb{R}^{N \times C}$ using a transposition operation and a linear embedding, where $C$ is the embedding dimension. 
Then, a learnable temporal position embedding $E^{m}_{TPos} {\in} \mathbb{R}^{N \times C}$ is utilized to retain positional information of frames, which can be formulated as: $\widetilde{Z}^{m}_{0} {=} \widetilde{Z}^{m} {+} E^{m}_{TPos}$. 

\subsection{Self-Hypothesis Refinement}
In the temporal domain, we first construct the SHR to refine single-hypothesis features. 
Each SHR layer consists of a multi-hypothesis self-attention (MH-SA) block and a hypothesis-mixing MLP block. 

\noindent \textbf{MH-SA.}
The core of the Transformer model is MSA, through which any two elements can interact with each other, thus modeling long-range dependencies. 
Instead, our MH-SA aims to capture single-hypothesis dependencies within each hypothesis independently for self-hypothesis communication. 
Specifically, the embedded features $\widetilde{Z}_{0}^{m} {\in} \mathbb{R}^{N \times C}$ of different hypotheses are fed into several parallel MSA blocks, which can be expressed as:
\begin{equation}
  \widetilde{Z}_{l}^{\prime m} = \widetilde{Z}_{l-1}^{m}+\operatorname{MSA}^{m}(\operatorname{LN}(\widetilde{Z}^{m}_{l-1})), 
\end{equation}
where $l {\in} [1, ..., L_{2}]$ is the index of SHR layers. 
Therefore, the message of different hypothesis features can be passed in a self-hypothesis way for feature enhancement. 

\noindent \textbf{Hypothesis-Mixing MLP.}
The multiple hypotheses are processed independently in the MH-SA, but there is no information exchange across hypotheses. 
To handle this issue, we add a hypothesis-mixing MLP after the MH-SA.
The features of multiple hypotheses are concatenated and fed into the hypothesis-mixing MLP to merge (\emph{i.e.}, converge) themselves. 
Then, the converged features are evenly partitioned (\emph{i.e.}, diverged) into non-overlapping chunks along the channel dimension, forming refined hypothesis representations. 
The procedure can be formulated as:
\begin{align}
  \begin{split}
  \label{equ:mmlp_smt}
  \widetilde{Z}^{\prime}_{l} {=} \operatorname{Concat}(\widetilde{Z}^{\prime 1}_{l}, ..., \widetilde{Z}^{\prime M}_{l})  \in \mathbb{R}^{N \times (C \cdot M)}, \\
  \operatorname{Concat}(\widetilde{Z}^{1}_{l}, ..., \widetilde{Z}^{M}_{l}) {=} \widetilde{Z}^{\prime}_{l} + \operatorname{HM-MLP}(\operatorname{LN}(\widetilde{Z}^{\prime}_{l})),
  \end{split}
\end{align}
where $\operatorname{Concat}(\cdot)$ is the concatenation operation and $\operatorname{HM-MLP}(\cdot)$ is the function of hypothesis-mixing MLP which shares the same format as Eq.~(\ref{equ:mlp}). 
This process explores the relations among channels of different hypotheses. 

\subsection{Cross-Hypothesis Interaction}
We then model interactions among multi-hypothesis features via the CHI, which contains two blocks: multi-hypothesis cross-attention (MH-CA) and hypothesis-mixing MLP. 

\begin{figure}[tb]
  \centering
  \includegraphics[width=1.00\linewidth]{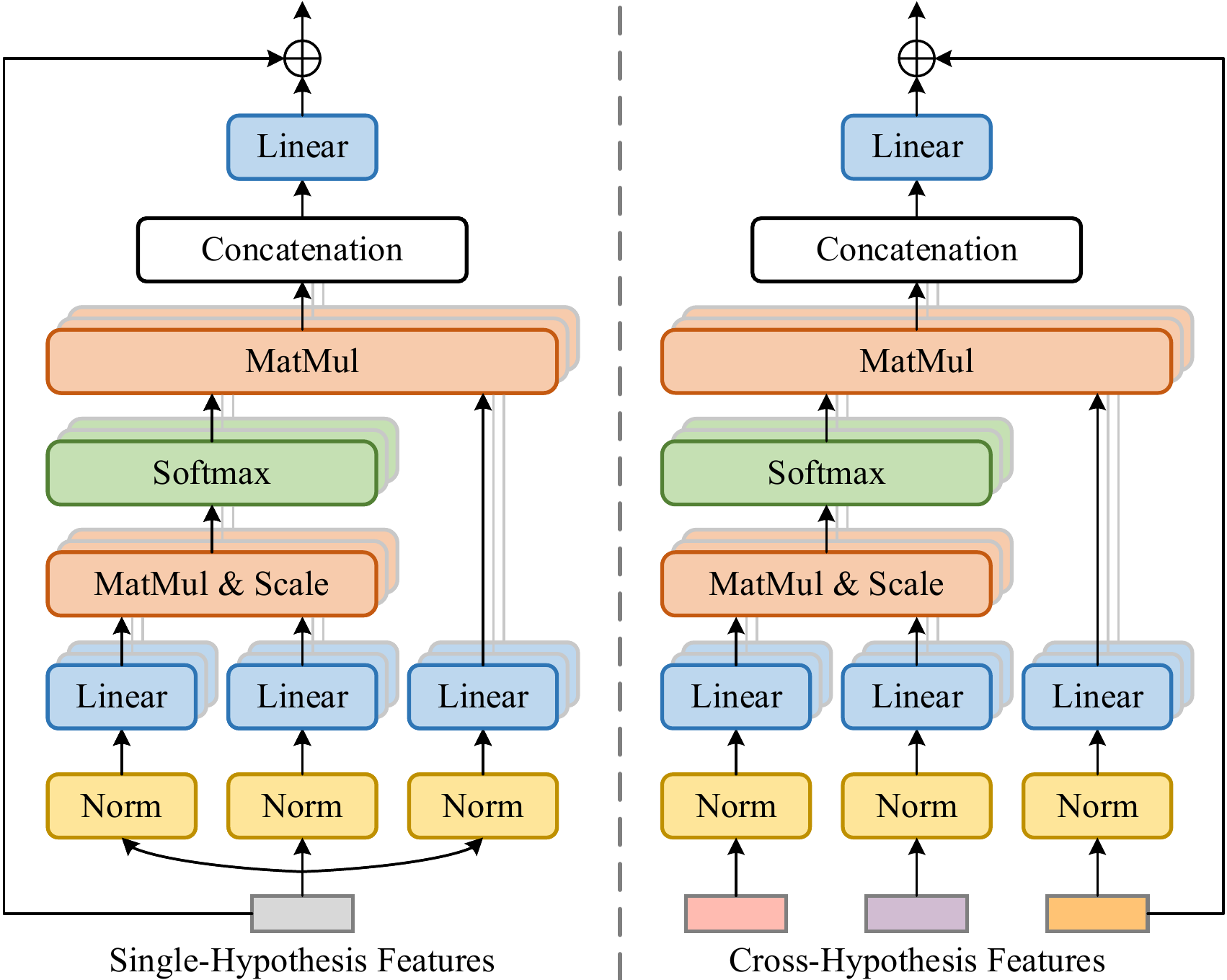}
  \caption
  {\textbf{Left}: Multi-head self-attention (MSA). 
  \textbf{Right}: Multi-head cross-attention (MCA). 
  }
  \label{fig:attention}
  \VspaceL
\end{figure}

\noindent \textbf{MH-CA.}
The MH-SA lacks connections across hypotheses, which limits its interaction modeling. 
To capture multi-hypothesis correlations mutually for cross-hypothesis communication, the MH-CA composed of multiple multi-head cross-attention (MCA) elements in parallel is proposed. 

The MCA measures the correlation among cross-hypothesis features and has a similar structure to MSA. 
The common configuration of MCA uses the same input between keys and values~\cite{liu2021video,xu2021cdtrans,chen2021crossvit}. 
However, an issue with this configuration is that it will result in more blocks (\textit{e.g.}, 6 MCA blocks for 3 hypotheses). 
Here, we adopt a more efficient strategy, which reduces the number of parameters by using different inputs (only require 3 MCA blocks), as shown in Figure~\ref{fig:attention} (Right). 
The multiple hypotheses $Z^{m}$ are alternately regarded as queries, keys, and values and fed into the MH-CA:
\begin{equation}
  {Z}_{l}^{\prime m} {=} {Z}_{l-1}^{m}{+}\operatorname{MCA}^{m}(\operatorname{LN}({Z}^{m_{1}}_{l-1}),\operatorname{LN}({Z}^{m_{2}}_{l-1}),\operatorname{LN}({Z}^{m}_{l-1})), \\
\end{equation}
where $l {\in} [1, ..., L_{3}]$ is the index of CHI layers, ${Z}_{0}^{m} {=} \widetilde{Z}^{m}_{L_{2}}$, $m_1$ and $m_2$ are the other two corresponding hypotheses, and $\operatorname{MCA}(Q,K,V)$ denotes the function of the MCA. 
Thanks to the MH-CA, the message passing can be performed in a crossing way to significantly improve modeling power. 

\noindent \textbf{Hypothesis-Mixing MLP.}
The hypothesis-mixing MLP in the CHI serves as the same function as the process in Eq.~(\ref{equ:mmlp_smt}). 
The outputs of the MH-CA are fed into it: 
\begin{align}
  \begin{split}
  {Z}^{\prime}_{l} {=}
  \operatorname{Concat}({Z}^{\prime 1}_{l}, ..., {Z}^{\prime M}_{l}) \in \mathbb{R}^{N \times (C \cdot M)}, \\
  \operatorname{Concat}({Z}^{1}_{l}, ..., {Z}^{M}_{l}) {=} {Z}^{\prime}_{l} + \operatorname{HM-MLP}(\operatorname{LN}({Z}^{\prime}_{l})).
  \end{split}
\end{align}

In the hypothesis-mixing MLP of the last CHI layer, the partition operation is not used so that the features of all hypotheses are finally aggregated to synthesize a single hypothesis representation ${Z}_{L_{3}} {\in} \mathbb{R}^{N \times (C \cdot M)}$. 

\subsection{Regression Head}
In the regression head, a linear transformation layer is applied on the output ${Z}_{L_{3}}$ to perform regression to produce the 3D pose sequence $\widetilde{X} {\in} \mathbb{R}^{N \times J \times 3}$. 
Finally, the 3D pose of the center frame $\hat{X} {\in} \mathbb{R}^{J \times 3}$ is selected from $\widetilde{X}$ as our final prediction. 

\subsection{Loss Function}
The entire model is trained in an end-to-end manner with a Mean Squared Error (MSE) loss,  
which is applied to minimize the error between the estimated and ground truth poses:
\begin{equation}
    \mathcal{L}=\sum_{n=1}^{N} \sum_{i=1}^{J}\left\|Y_{i}^{n}-\widetilde{X}_{i}^{n}\right\|_{2},
\end{equation}
where $\widetilde{X}_{i}^{n}$ and $Y_{i}^{n}$ represent the predicted and ground truth 3D poses of joint $i$ at frame $n$, respectively. 

\begin{table*}[htb]
  \normalsize
  \centering
  \caption
  {
    Quantitative comparison with the state-of-the-art methods on Human3.6M under Protocol 1, using detected 2D poses (top) and ground truth 2D poses (bottom) as inputs. 
    $(\dagger)$ - uses temporal information. 
    \textbf{Blod}: best; 
    \underline{Underlined}: second best.  
  } 
  \resizebox{\textwidth}{!}{
  \begin{tabular}{@{}l|ccccccccccccccc|c@{}}
  \toprule[1pt]
  Method & Dir. & Disc & Eat & Greet & Phone & Photo & Pose & Purch. & Sit & SitD. & Smoke & Wait & WalkD. & Walk & WalkT. & Avg.\\
  \midrule[0.5pt]

  Fang \emph{et al.} (AAAI'18)~\cite{fang2018learning} & 50.1& 54.3& 57.0& 57.1& 66.6& 73.3& 53.4& 55.7& 72.8& 88.6& 60.3& 57.7& 62.7& 47.5& 50.6& 60.4 \\

  GraphSH (CVPR'21)~\cite{xu2021graph} &45.2 &49.9 &47.5 &50.9 &54.9 &66.1 &48.5 &46.3 &59.7 &71.5 &51.4 &48.6 &53.9 &39.9 &44.1 &51.9 \\

  MGCN (ICCV'21)~\cite{zou2021modulated} &45.4 &49.2 &45.7 &49.4 &50.4 &58.2 &47.9 &46.0 &57.5 &63.0 &49.7 &46.6 &52.2 &38.9 &40.8 &49.4 \\

  ST-GCN (ICCV'19)~\cite{cai2019exploiting} $(\dagger)$ &44.6 &47.4 &45.6 &48.8 &50.8 &59.0 &47.2 &43.9&57.9 &61.9 &49.7 &46.6 &51.3 &37.1 &39.4 &48.8 \\
  
  VPose (CVPR'19)~\cite{pavllo20193d} $(\dagger)$ & 45.2 & 46.7 & 43.3 & 45.6 & 48.1 & 55.1 & 44.6 & 44.3 & 57.3 & 65.8 & 47.1 & 44.0 & 49.0 & 32.8 & 33.9 & 46.8 \\

  SGNN (ICCV'21)~\cite{zeng2021learning} $(\dagger)$ &- &- &- &- &- &- &- &- &- &- &- &- &- &- &- &45.7 \\

  UGCN (ECCV'20)~\cite{wang2020motion} $(\dagger)$ &\underline{41.3} &43.9 &44.0 &\underline{42.2} &48.0 &57.1 &42.2 &43.2 &57.3 &61.3 &47.0 &43.5 &47.0 &32.6 &\underline{31.8} &45.6 \\

  Liu \emph{et al.} (CVPR'20)~\cite{liu2020attention} $(\dagger)$ &41.8 &44.8 &41.1 &44.9 &47.4 &54.1 &43.4 &42.2 &56.2 &63.6 &\underline{45.3} &43.5 &\underline{45.3} &\underline{31.3} &32.2 &45.1 \\

  PoseFormer (ICCV'21)~\cite{poseformer} $(\dagger)$ &{41.5} &44.8 &\textbf{39.8} &42.5 &\underline{46.5} &\underline{51.6} &42.1 &\underline{42.0} &\textbf{53.3} &{60.7} &45.5 &43.3 &46.1 &31.8 &32.2 &44.3 \\

  Anatomy3D (TCSVT'21)~\cite{chen2021anatomy} $(\dagger)$ &41.4 &\underline{43.2} &\underline{40.1} &42.9 &46.6 &{51.9} &\underline{41.7} &42.3 &53.9 &\textbf{60.2} &45.4 &\underline{41.7} &46.0 &31.5 &32.7 &\underline{44.1} \\
  
  \midrule[0.5pt]

  MHFormer (Ours) $(\dagger)$ &\textbf{39.2} &\textbf{43.1} &\underline{40.1} &\textbf{40.9} &\textbf{44.9} &\textbf{51.2} &\textbf{40.6} &\textbf{41.3} &\underline{53.5} &\underline{60.3} &\textbf{43.7} &\textbf{41.1} &\textbf{43.8} &\textbf{29.8} &\textbf{30.6} &\textbf{43.0} \\

  \toprule[1pt]
  Method & Dir. & Disc & Eat & Greet & Phone & Photo & Pose & Purch. & Sit & SitD. & Smoke & Wait & WalkD. & Walk & WalkT. & Avg.\\
  \midrule[0.5pt]

  P-LSTM (ECCV'18)~\cite{lee2018propagating} $(\dagger)$ &32.1 &36.6 &34.3 &37.8 &44.5 &49.9 &40.9 &36.2 &44.1 &45.6 &35.3 &35.9 &30.3 &37.6 &35.5 &38.4 \\

  PoseAug (CVPR'21)~\cite{gong2021poseaug}  &- &- &- &- &- &- &- &- &- &- &- &- &- &- &-  &38.2 \\

  VPose (CVPR'19)~\cite{pavllo20193d} $(\dagger)$ &35.2 &40.2 &32.7 &35.7 &38.2 &45.5 &40.6 &36.1 &48.8 &47.3 &37.8 &39.7 &38.7 &27.8 & 29.5 &37.8 \\

  Liu \emph{et al.} (CVPR'20)~\cite{liu2020attention} $(\dagger)$ &34.5 &37.1 &33.6 &34.2 &32.9 &37.1 &39.6 &35.8 &40.7 &41.4 &33.0 &33.8 &33.0 &26.6 &26.9 &34.7 \\

  Anatomy3D (TCSVT'21)~\cite{chen2021anatomy} $(\dagger)$ &- &- &- &- &- &- &- &- &- &- &- &- &- &- &- &32.3 \\
    
  SRNet (ECCV'20)~\cite{zeng2020srnet} $(\dagger)$ &34.8 &\textbf{32.1} &\textbf{28.5} &\underline{30.7} &{31.4} &{36.9} &35.6 &\textbf{30.5} &{38.9} &40.5 &32.5 &\textbf{31.0} &29.9 &\underline{22.5} &24.5 &32.0 \\

  PoseFormer (ICCV'21)~\cite{poseformer} $(\dagger)$ &\underline{30.0} &\underline{33.6} &{29.9} &31.0 &\underline{30.2} &\textbf{33.3} &\underline{34.8} &31.4 &\underline{37.8} &\textbf{38.6} &\underline{31.7} &\underline{31.5} &\textbf{29.0} &23.3 &\underline{23.1} &\underline{31.3} \\

  \midrule[0.5pt]
  
  MHFormer (Ours) $(\dagger)$ &\textbf{27.7} &\textbf{32.1} &\underline{29.1} &\textbf{28.9} &\textbf{30.0} &\underline{33.9} &\textbf{33.0} &\underline{31.2} &\textbf{37.0} &\underline{39.3} &\textbf{30.0} &\textbf{31.0} &\underline{29.4} &\textbf{22.2} &\textbf{23.0} &\textbf{30.5}

 \\
  \toprule[1pt]
  \end{tabular}
  }
  \VspaceS
  \label{table:h36m}
\end{table*}

\begin{table}[tb]
  \centering  
  \footnotesize
  \caption
  {
    Comparison with the methods of generating multiple 3D pose hypotheses on Human3.6M. 
    The number of hypotheses is denoted as $M$. 
    \textbf{Blod}: best; 
    \underline{Underlined}: second best. 
  }
  \setlength{\tabcolsep}{4.20mm}  
  \begin{tabular}{lcc}
  \toprule [1pt]
  Method &$M$ &MPJPE ($mm$) \\
  \midrule  [0.5pt]
  Li \emph{et al.} (CVPR'19)~\cite{li2019generating} &5 &52.7 \\
  Sharma \emph{et al.} (ICCV'19)~\cite{sharma2019monocular} &200 &46.8 \\
  Oikarinen (IJCNN'21)~\cite{oikarinen2021graphmdn} &200 &46.2 \\
  Wehrbein \emph{et al.} (ICCV'21)~\cite{wehrbein2021probabilistic} &200 &\underline{44.3} \\
  \midrule
  MHFormer (Ours) &3 &\textbf{43.0} \\
  \toprule [1pt]
  \end{tabular}
  \label{table:hypotheses}
  \VspaceL
\end{table}

\section{Experiments}
\subsection{Datasets and Evaluation Metrics}
We evaluate our method on two widely-used datasets for 3D HPE: Human3.6M~\cite{ionescu2013human3} and MPI-INF-3DHP~\cite{mehta2017monocular}. 

\noindent \textbf{Human3.6M.}
The Human3.6M dataset~\cite{ionescu2013human3} is the largest and most representative benchmark for 3D HPE. 
This dataset consists of 3.6 million images captured from four synchronized cameras at 50 Hz. 
There are 15 daily activities performed by 11 human subjects in an indoor environment. 
Following previous works~\cite{pavllo20193d,liu2020attention,wang2020motion,chen2021anatomy}, we train a single model on five subjects (S1, S5, S6, S7, S8) and test it on two subjects (S9 and S11). 
We adopt the most commonly used evaluation protocols: 
Protocol 1 is the MPJPE which measures the mean Euclidean distance between the ground truth and estimated joints in millimeters; 
Protocol 2 is the MPJPE after aligning the predicted 3D pose with the ground truth using translation, rotation, and scale (P-MPJPE). 

\noindent \textbf{MPI-INF-3DHP.}
The MPI-INF-3DHP~\cite{mehta2017monocular} is a large 3D pose dataset in both indoor and outdoor environments. 
This dataset provides 1.3 million frames, containing more diverse motions than Human3.6M. 
Following the setting in~\cite{mehta2017monocular,lin2019trajectory,chen2021anatomy,poseformer}, we report metrics of MPJPE, Percentage of Correct Keypoint (PCK) with the threshold of 150 $mm$, and Area Under Curve (AUC) for a range of PCK thresholds. 

\subsection{Implementation Details}
In our implementation, the proposed MHFormer contains $L_{1} {=} 4$ MHG, $L_{2} {=} 2$ SHR, and $L_{3} {=} 1$ CHI layers. 
The MHFormer model is implemented in PyTorch framework on one GeForce RTX 3090 GPU. 
We train our model in an end-to-end manner from scratch using Amsgrad optimizer. 
The initial learning rate is set to 0.001 with a shrink factor of 0.95 applied after each epoch and 0.5 after every 5 epochs. 
For a fair comparison, the same horizontal flip augmentation is adopted following~\cite{pavllo20193d,cai2019exploiting,chen2021anatomy,poseformer}. 
We perform the 2D pose detection using cascaded pyramid network (CPN)~\cite{chen2018cascaded} for Human3.6M following~\cite{pavllo20193d,cai2019exploiting,liu2020attention} and ground truth 2D pose for MPI-INF-3DHP following~\cite{lin2019trajectory,chen2021anatomy,poseformer}. 

\begin{figure*}[htb]
  \centering
  \includegraphics[width=1.00\linewidth]{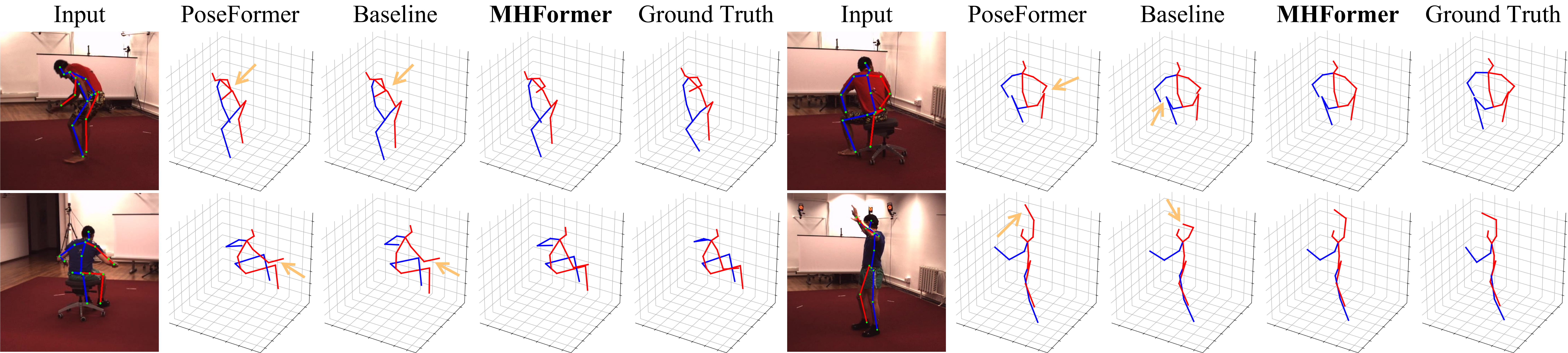}
  \caption
  {Qualitative comparison among the proposed method (MHFormer), the baseline method, and the previous state-of-the-art method (PoseFormer)~\cite{poseformer} on Human3.6M dataset. 
  Wrong estimations are highlighted by yellow arrows. 
  }
  \label{fig:result}
  \VspaceS
\end{figure*}

\subsection{Comparison with State-of-the-Art Methods}
\noindent
\textbf{Results on Human3.6M.}
The proposed MHFormer is compared with the state-of-the-art methods on Human3.6M. 
The results of our model with a receptive field of 351 frames using 2D detected inputs~\cite{chen2018cascaded} are reported in Table~\ref {table:h36m} (top). 
Without bells and whistles, our MHFormer outperforms all previous state-of-the-art methods by a large margin under both Protocol 1 (43.0 $mm$) and Protocol 2 (34.4 $mm$, see supplemental material). 
Compared to the very recent Transformer-based method, \emph{i.e.}, PoseFormer~\cite{poseformer}, MHFormer noticeably surpasses it by 1.3 $mm$ in MPJPE (relative 3\% improvement). 
Figure~\ref{fig:result} shows the qualitative comparison with the PoseFormer and the baseline model (same architecture as ViT~\cite{dosovitskiy2020image}) on some challenging poses. 
To further explore the lower bound of our method, we compared our MHFormer with the state-of-the-art methods with ground truth 2D poses as inputs. 
The results are shown in Table~\ref{table:h36m} (bottom). 
It can be seen that our method achieves the best performance (30.5 $mm$ in MPJPE), outperforming all other methods. 

Additionally, our method is compared with previous methods of generating multiple 3D pose hypotheses. 
The results are shown in Table~\ref{table:hypotheses}. 
It is noteworthy that these methods report metrics for the best hypothesis due to the adopted one-to-many mapping, while our method reports metrics with a specific solution by learning a deterministic mapping, which is much more practical in reality. 
Even though we use much fewer hypothesis numbers (3 vs. 200), our proposed method consistently outperforms previous works. 

\begin{table}
  \footnotesize
  \centering
  \caption
  {
    Quantitative comparison with the state-of-the-art methods on MPI-INF-3DHP. 
    Best in bold, second best underlined.
  }
  \setlength{\tabcolsep}{3.90mm} 
  \begin{tabular}{@{}l|ccc@{}}
  \toprule
  Method & PCK $\uparrow$ & AUC $\uparrow$ & MPJPE $\downarrow$ \\
  \midrule
  Mehta \emph{et al.} (3DV'17)~\cite{mehta2017monocular} & 75.7 & 39.3 & 117.6 \\
  Lin \emph{et al.} (BMVC'19)~\cite{lin2019trajectory} & 83.6 & 51.4 & 79.8 \\
  VPose (CVPR'19)~\cite{pavllo20193d} &86.0 &51.9 &84.0 \\
  Li \emph{et al.} (CVPR'20)~\cite{li2020cascaded} &81.2 &46.1 &99.7 \\
  Anatomy3D (TCSVT'21)~\cite{chen2021anatomy} &87.9 &54.0 &78.8 \\
  PoseFormer (ICCV'21)~\cite{poseformer} &\underline{88.6} &\underline{56.4} &\underline{77.1} \\ 

  \midrule
  MHFormer (Ours) &\textbf{93.8} &\textbf{63.3} &\textbf{58.0} \\
  \bottomrule
  \end{tabular}
  \label{table:3dhp}
  \VspaceS
\end{table}

\begin{table}[tb]
  \centering  
  \footnotesize
  \caption
  {
    Ablation study on different receptive fields with MPJPE ($mm$).
    CPN - cascaded pyramid network; 
    GT - 2D ground truth.
  }
  \setlength{\tabcolsep}{4.25mm}
  \begin{tabular}{cccccc}
  \toprule [1pt]
   &9 &27 &81 &243 &351 \\
  \midrule [0.5pt]
  CPN &47.8 &45.9 &44.5 &43.2 &\textbf{43.0} \\
  GT &36.6 &34.3 &32.7 &30.9 &\textbf{30.5} \\
  \toprule [1pt]
  \end{tabular}
  \label{table:frames}
  \VspaceL
\end{table}

\noindent \textbf{Results on MPI-INF-3DHP.}
To assess the generalization ability, we evaluate our method on MPI-INF-3DHP dataset. 
Following~\cite{poseformer}, we use 2D pose sequences of 9 frames as our model input due to the fewer samples and shorter sequence lengths of this dataset compared to Human3.6M. 
The results in Table~\ref{table:3dhp} demonstrate that our method achieves the best performance on all metrics (PCK, AUC, and MPJPE). 
It emphasizes the effectiveness of our MHFormer in improving performance in outdoor scenes. 

\subsection{Ablation Study}
To verify the impact of each component and design in the proposed model, we conduct 
extensive ablation experiments on Human3.6M dataset under Protocol 1 with MPJPE. 

\noindent
\textbf{Impact of Receptive Fields.}
For the video-based 3D HPE task, a large receptive field is essential for estimation accuracy. 
Table~\ref{table:frames} shows the results of our method with different input frames. 
It can be seen that our method obtains larger gains with more frames fed into the model.  
The error has a great decrease of 16.7\% from 9-frames to 351-frames with GT input, which indicates the effectiveness of our method in capturing long-range dependencies across frames with a large receptive field. 
Next, ablations in the following parts are carried out using a receptive field of 27 frames to balance the computation efficiency and performance.

\begin{table}[tb]
  \centering  
  \footnotesize
  \caption
  {
    Ablation study on different parameters of MHG. 
    Here, $L_{1}$ is the number of MHG layers and $M$ is the hypothesis number. 
  }
  \setlength{\tabcolsep}{3.30mm}  
  \begin{tabular}{ccccc}
    \toprule [1pt]
    $M$ &$L_{1}$  &Params (M) &FLOPs (G) &MPJPE ($mm$) \\
    \midrule [0.5pt]
    3& 2& 18.91& 1.03& 46.4 \\
    3& 3& 18.92& 1.03& 46.3 \\
    3& 4& 18.92& 1.03& \textbf{45.9} \\
    3& 5& 18.93& 1.04& 46.1 \\
    \midrule [0.5pt]
    1& 4& 6.32& 0.34& 47.6 \\
    2& 4& 12.61& 0.69& 46.7 \\
    3& 4& 18.92& 1.03& \textbf{45.9} \\
    4& 4& 25.22& 1.38& 46.9 \\
    \toprule [1pt]
  \end{tabular}
  \label{table:MHG}
  \VspaceL
\end{table}

\begin{figure*}[htb]
  \centering
  \includegraphics[width=1.00\linewidth]{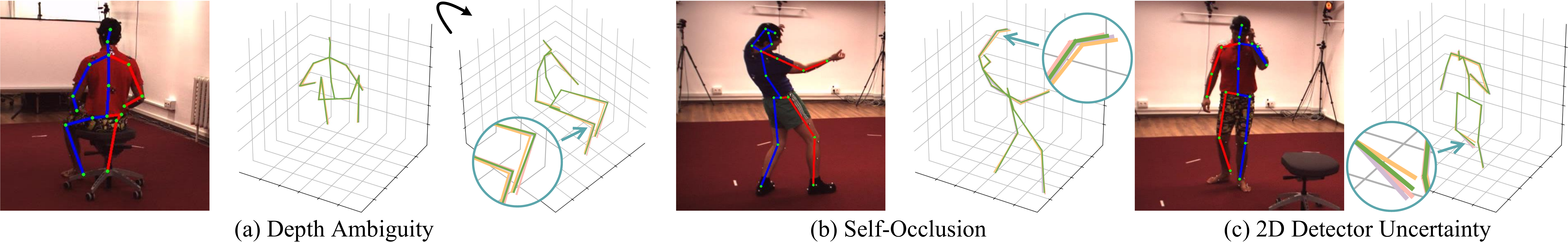}
  \caption
  {Diverse 3D pose hypotheses generated by MHFormer. 
  For easy illustration, we color-code the hypotheses to show the difference among them. 
  Green colored 3D pose corresponds to the final synthesized estimation of our method. 
  }
  \label{fig:hypotheses}
  \VspaceS
\end{figure*}

\noindent
\textbf{Impact of Parameters in MHG.}
In the top part of Table~\ref{table:MHG}, we report the results with different numbers of MHG layers. 
Experiments show that stacking more layers in MHG can slightly improve the performance with few parameter increases, but the gain disappears when the layer number is larger than 4. 
Moreover, we investigate the influence of using different numbers of hypotheses in MHG. 
The results are shown in the bottom part of Table~\ref{table:MHG}. 
Increasing the number of hypotheses can improve the result, but the performance saturates when using 3 hypothesis representations. 
Notably, our model equipped with 3 hypotheses shows significant gains over the single-hypothesis model with 1.7 $mm$ error reduction. 
This demonstrates that exploiting different representations of multiple pose hypotheses is helpful to improve the performance of the model, validating our motivation. 

\noindent \textbf{Impact of Parameters in SHR and CHI.}
Table~\ref{table:SHRandCHI} reports how the different parameters of SHR and CHI impact the performance and computation complexity of our model. 
The results show that enlarging the embedding dimension from 256 to 512 can boost the performance, but using dimensions larger than 512 cannot bring further improvements. 
In addition, we observe no more gains by stacking either more SHR or CHI layers. 
Therefore, the optimal parameters for our model are $L_{2}{=}2$, $L_{3}{=}1$, and $C{=}512$. 

\begin{table}[tb]
  \centering  
  \footnotesize
  \caption
  {
    Ablation study on different parameters of SHR and CHI.
    Here, $L_{2}$ and $L_{3}$ indicate the number of SHR and CHI layers, respectively. 
    $C$ is the embedding dimension. 
  }
  \setlength{\tabcolsep}{2.35mm}  
  \begin{tabular}{cccccc}
  \toprule  [1pt]
  $L_{2}$& $L_{3}$& $C$ & Params (M) &FLOPs (G) &MPJPE ($mm$) \\
  \midrule [0.5pt]
  2& 1& 256& 4.72& 0.26& 47.2 \\
  2& 1& 384& 10.65& 0.58& 46.4 \\
  2& 1& 512& 18.92& 1.03& \textbf{45.9} \\
  2& 1& 768& 42.50& 2.31& 47.4 \\
  \midrule [0.5pt]

  2& 1& 512& 18.92& 1.03& \textbf{45.9} \\
  1& 3& 512& 25.20& 1.38&46.7 \\
  2& 2& 512& 25.20& 1.38& 46.8 \\
  3& 1& 512& 25.20& 1.38& 46.3 \\
  \toprule [1pt]
  \end{tabular}
  \label{table:SHRandCHI}
  \VspaceL
\end{table}

\noindent
\textbf{Effect of Model Components.}
In Table~\ref{table:component}, we carry out experiments to quantify the influence of our proposed components. 
Firstly, we compare our method with the baseline model. 
For a fair comparison, the results of the baseline are reported at the same number of layers as MHFormer with different embedding dimensions, since our hypothesis-mixing MLP in MHFormer takes concatenated hypothesis features as inputs (the dimension is $512 {\times} 3 {=} 1536$).
The results show that the baseline model is prone to overfitting due to the excessive number of parameters, whereas our method performs well. 
Additionally, it can be seen that our MHFormer built upon MHG, SHR, and CHI outperforms varying variants of baseline models (1.9 $mm$ improvement). 
Then, when we incorporate multi-hypothesis representations and SHR or CHI within the baseline, the performance has significant gains (-1.3 $mm$ for MHG-SHR and -1.0 $mm$ for MHG-CHI). 
Besides, we remove the MHG in MHFormer (SHR-CHI). 
At this point, the model only captures temporal information and its error heavily increases by 1.3 $mm$. 
These ablations indicate that learning multi-hypothesis spatio-temporal representations is significant for 3D HPE, and the different hypothesis representations should be modeled in both independent and mutual ways. 

We also explore the use of multi-level features in MHG by simply building the MHG upon several parallel Transformer encoders (MHFormer $^{*}$). 
As shown in the table, our MHFormer equipped with the multi-level features increases the performance, which indicates that the multi-level features can bring valuable information to the final estimation. 

\begin{table}[t]
  \centering
  \footnotesize
  \caption
  {
    Ablation study on different components of our MHFormer. 
    Here, $^{*}$ means no multi-level features in MHG. 
  }
  \setlength{\tabcolsep}{2.50mm}
  \begin{tabular}{l|ccc|c}
  \toprule  [1pt]
  Method &MHG &SHR &CHI &MPJPE ($mm$) \\
  \midrule [0.5pt]
  Baseline ($C{=}256$) &\xmark &\xmark &\xmark &49.9 \\
  Baseline ($C{=}512$) &\xmark &\xmark &\xmark &47.8 \\
  Baseline ($C{=}1536$) &\xmark &\xmark &\xmark &50.4 \\
  \midrule [0.5pt]

  SHR-CHI &\xmark &\cmark &\cmark &47.2 \\
  MHG-SHR &\cmark &\cmark &\xmark  &46.5 \\
  MHG-CHI &\cmark &\xmark &\cmark &46.8 \\
  MHFormer $^{*}$ &\cmark &\cmark &\cmark &46.5 \\
  MHFormer (Ours) &\cmark &\cmark &\cmark &\textbf{45.9} \\
  \toprule [1pt]
  \end{tabular}
  \label{table:component}
  \VspaceL
\end{table}

\section{Qualitative Results}
Although our method does not aim to produce multiple 3D pose predictions, for better observation, we add additional regression layers and finetune our model to visualize the intermediate hypotheses.
The several qualitative results are shown in Figure~\ref{fig:hypotheses}. 
It can be seen that our method is able to generate different plausible 3D pose solutions, especially for ambiguous body parts with depth ambiguity, self-occlusion, and 2D detector uncertainty. 
Moreover, the final 3D pose synthesized by aggregating multi-hypothesis information is more reasonable and accurate. 

\section{Conclusion}
This paper presents Multi-Hypothesis Transformer (MHFormer), a new Transformer-based three-stage framework for the ambiguous inverse problem of 3D HPE from monocular videos. 
MHFormer first generates initial representations of multiple pose hypotheses in the spatial domain and then communicates across them in both independent and mutual ways in the temporal domain. 
Extensive experiments show that the proposed MHFormer has a fundamental advantage over single-hypothesis Transformers and achieves state-of-the-art performance on two benchmark datasets. 
We hope that our approach will foster further research in 2D-to-3D pose lifting considering various ambiguities. 

\noindent \textbf{Limitation.}
One limitation of our method is the relatively larger computational complexity.
The excellent performance of Transformers comes at a price of high computational cost. 

{\small
\bibliographystyle{ieee_fullname}
\bibliography{ref}
}

\newpage
\appendix
{\noindent\Large\textbf{Supplementary Material}}
\newline

This supplementary material contains the following details:
(1) A brief description of multi-head cross-attention. 
(2) Additional quantitative results. 
(3) Additional ablation studies.  
(4) Additional visualization results. 

\section{Multi-Head Cross-Attention}
In Sec. 3.1 of our main manuscript, we give a brief description of the multi-head self-attention (MSA) block. 
Given the inputs $x {\in} \mathbb{R}^{n \times d}$, they are linearly mapped to queries $Q {\in} \mathbb{R}^{n \times d}$, keys $K {\in} \mathbb{R}^{n \times d}$, and values $V {\in} \mathbb{R}^{n \times d}$. 
The scaled dot-product attention in the MSA can be computed by:
\begin{equation}
   \operatorname{Attention}(Q, K, V)=\operatorname{Softmax}\left(\frac{{Q K^{T}}} {{\sqrt{d}}}\right) V.
\end{equation}

In this section, we further define the multi-head cross-attention (MCA) among three tensors, $x$, $y$, and $z$. 
The inputs $x {\in} \mathbb{R}^{n \times d}$, $y {\in} \mathbb{R}^{n \times d}$, and $z {\in} \mathbb{R}^{n \times d}$ are linearly mapped to queries $Q_{x} {\in} \mathbb{R}^{n \times d}$, keys $K_{y} {\in} \mathbb{R}^{n \times d}$, and values $V_{z} {\in} \mathbb{R}^{n \times d}$, respectively. 
The scaled dot-product attention in the MCA can be computed by:    
\begin{equation}
   \operatorname{Attention_{cross}}(Q_{x}, K_{y}, V_{z})=\operatorname{Softmax}\left(\frac{{Q_{x} K_{y}^{T}}} {{\sqrt{d}}}\right) V_{z}.
\end{equation}
The common configuration of MCA uses the same input between keys and values~\cite{liu2021video,xu2021cdtrans,chen2021crossvit}, \textit{i.e.}, the inputs $x \neq y = z$. 
Instead, we adopt a more efficient strategy by using different inputs, \textit{i.e.}, the inputs $x \neq y \neq z$. 

\section{Additional Quantitative Results}
Table~\ref{table:h36m} shows the results of our proposed MHFormer on Human3.6M under Protocol 2. 
The input 2D poses are estimated by CPN~\cite{chen2018cascaded}.
Without bells and whistles, our MHFormer achieves promising results that outperform the state-of-the-art approaches. 

Several methods~\cite{cai2019exploiting,wang2020motion,zou2021modulated} adopt a pose refinement module, which is first proposed by ST-GCN~\cite{cai2019exploiting}, to further improve the estimation accuracy. 
Following~\cite{cai2019exploiting}, we adopt the refine module and the results are shown in Table~\ref{table:refine}. 
It can be seen that our method can use the refine module to improve the performance, achieving an error of 42.4 $mm$ in MPJPE which surpasses all other approaches by a large margin. 

\begin{table*}[htb]
\normalsize
\centering
\caption
{
   Quantitative comparison with the state-of-the-art methods on Human3.6M under Protocol 2. 
   $(\dagger)$ - uses temporal information. 
   \textbf{Blod}: best; 
   \underline{Underlined}: second best. 
} 
\resizebox{\textwidth}{!}{
\begin{tabular}{@{}l|ccccccccccccccc|c@{}}
\toprule[1pt]
Method & Dir. & Disc & Eat & Greet & Phone & Photo & Pose & Purch. & Sit & SitD. & Smoke & Wait & WalkD. & Walk & WalkT. & Avg.\\
\midrule[0.5pt]

SimpleBaseline (ICCV'17)~\cite{martinez2017simple} &39.5 &43.2 &46.4 &47.0 &51.0 &56.0 &41.4 &40.6 &56.5 &69.4 &49.2 &45.0 &49.5 &38.0 &43.1 &47.7 \\

Fang \emph{et al.} (AAAI'18)~\cite{fang2018learning} &38.2 &41.7 &43.7 &44.9 &48.5 &55.3 &40.2 &38.2 &54.5 &64.4 &47.2 &44.3 &47.3 &36.7 &41.7 &45.7 \\

PoseAug (CVPR'21)~\cite{gong2021poseaug}  &- &- &- &- &- &- &- &- &- &- &- &- &- &- &-  &39.1 \\

SGNN (ICCV'21)~\cite{zeng2021learning} &33.9 &37.2 &36.8 &38.1 &38.7 &43.5 &37.8 &35.0 &47.2 &53.8 &40.7 &38.3 &41.8 &30.1 &31.4 &39.0 \\

ST-GCN (ICCV'19)~\cite{cai2019exploiting} $(\dagger)$ &35.7 &37.8 &36.9 &40.7 &39.6 &45.2 &37.4 &34.5 &46.9 &50.1 &40.5 &36.1 &41.0 &29.6 &33.2 &39.0 \\

VPose \emph{et al.} (CVPR'19)~\cite{pavllo20193d} $(\dagger)$ &34.1 &36.1 &34.4 &37.2 &36.4 &42.2 &34.4 &33.6 &45.0 &52.5 &37.4 &33.8 &37.8 &25.6 &27.3 &36.5 \\

Xu \emph{et al.} (CVPR'20)~\cite{xu2020deep} $(\dagger)$ &\textbf{31.0} &\textbf{34.8} &34.7 &\underline{34.4} &36.2 &43.9 &\underline{31.6} &33.5 &\textbf{42.3} &{49.0} &37.1 &33.0 &39.1 &26.9 &31.9 &36.2 \\

Liu \emph{et al.} (CVPR'20)~\cite{liu2020attention} $(\dagger)$ &32.3 &35.2 &33.3 &35.8 &\underline{35.9} &41.5 &33.2 &32.7 &44.6 &50.9 &37.0 &\textbf{32.4} &37.0 &25.2 &27.2 &35.6 \\

UGCN (ECCV'20)~\cite{wang2020motion} $(\dagger)$ &32.9 &35.2 &35.6 &\underline{34.4} &36.4 &42.7 &\textbf{31.2} &32.5 &45.6 &50.2 &37.3 &32.8 &36.3 &26.0 &\textbf{23.9} &35.5 \\

Anatomy3D (TCSVT'21)~\cite{chen2021anatomy} $(\dagger)$ &32.6 &{35.1} &\underline{32.8} &35.4 &36.3 &{40.4} &32.4 &{32.3} &\underline{42.7} &{49.0} &36.8 &\textbf{32.4} &36.0 &24.9 &26.5 &35.0 \\

PoseFormer (ICCV'21)~\cite{poseformer} $(\dagger)$ &32.5 &\textbf{34.8} &\textbf{32.6} &34.6 &\textbf{35.3} &\textbf{39.5} &{32.1} &\textbf{32.0} &42.8 &\textbf{48.5} &\textbf{34.8} &\textbf{32.4} &\underline{35.3} &\underline{24.5} &26.0 &\underline{34.6} \\

\midrule[0.5pt]
MHFormer (Ours) $(\dagger)$ &\underline{31.5} &\underline{34.9} &\underline{32.8} &\textbf{33.6} &\textbf{35.3} &\underline{39.6} &{32.0} &\underline{32.2} &{43.5} &\underline{48.7} &\underline{36.4} &\underline{32.6} &\textbf{34.3} &\textbf{23.9} &\underline{25.1} &\textbf{34.4} \\

\toprule[1pt]
\end{tabular}
}
\VspaceS
\label{table:h36m}
\end{table*} 

\begin{table}
   \footnotesize
   \centering
   \caption
   {
     Quantitative comparison on Human3.6M under MPJPE. 
     \textbf{Blod}: best; 
     \underline{Underlined}: second best. 
   }
   \setlength{\tabcolsep}{4.40mm} 
   \begin{tabular}{@{}lcccc@{}}
   \toprule
   Method &Refine module &MPJPE ($mm$) \\
   \midrule
   MGCN (ICCV'21)~\cite{zou2021modulated} &\cmark &49.4 \\
   ST-GCN (ICCV'19)~\cite{cai2019exploiting} &\cmark &48.8 \\
   UGCN (ECCV'20)~\cite{wang2020motion} & &45.6 \\
   UGCN (ECCV'20)~\cite{wang2020motion} &\cmark &44.5 \\
   \midrule
   MHFormer (Ours) & &\underline{43.0} \\
   MHFormer (Ours) &\cmark &\textbf{42.4} \\
   \bottomrule
   \end{tabular}
   \VspaceS
   \label{table:refine}
\end{table}

\begin{table}[tb]
   \centering  
   \footnotesize
   \caption
   {
     Ablation study on different configurations of MH-CA on Human3.6M under MPJPE. 
     Here, $^{*}$ means using the same input between keys and values in MH-CA. 
   }
   \setlength{\tabcolsep}{3.55mm}  
   \begin{tabular}{lccc}
     \toprule [1pt]
     Method &Params (M) &FLOPs (G) &MPJPE ($mm$) \\
     \midrule [0.5pt]
     MH-CA $^{*}$ &22.07 &1.21 &46.1 \\
     MH-CA &18.92 &1.03 &\textbf{45.9} \\
     \toprule [1pt]
   \end{tabular}
   \VspaceL
   \label{table:MHCA}
\end{table}
 
\section{Additional Ablation Studies}
\noindent
\textbf{Effect of Model Components.}
Here, we give more details about how to build the different variants of MHFormer in Table 7 of our main manuscript:
\begin{itemize}
   \item {Baseline}: The baseline model contains 3 layers for standard Transformer encoder (same architecture as ViT~\cite{dosovitskiy2020image}). 
   \item {SHR-CHI}: We remove the MHG module. SHR-CHI contains $L_{2} {=} 2$ SHR and $L_{3} {=} 1$ CHI layers. 
   \item {MHG-SHR}: We replace the CHI layers in MHFormer with SHR layers. MHG-SHR contains $L_{1} {=} 4$ MHG and $L_{3} {=} 3$ SHR layers. 
   \item {MHG-CHI}: We replace the SHR layers in MHFormer with CHI layers. SHR-CHI contains $L_{1} {=} 4$ MHG and $L_{3} {=} 3$ CHI layers. 
   \item {MHFormer $^{*}$}: The MHG in MHFormer is simply built upon several parallel Transformer encoders. 
   \item {MHFormer}: Our proposed method that contains $L_{1} {=} 4$ MHG, $L_{2} {=} 2$ SHR, and $L_{3} {=} 1$ CHI layers. Please refer to Figure 3 in our main manuscript. 
\end{itemize}

\noindent
\textbf{Impact of Configurations in MH-CA.}
As mentioned in Sec. 3.5 of our main manuscript, the common configuration of MCA uses the same input between keys and values~\cite{liu2021video,xu2021cdtrans,chen2021crossvit}, which will result in more blocks. 
We adopt a more efficient configuration by using different inputs among queries, keys, and values. 
The performance and computational complexity of these two configurations are given in Table~\ref{table:MHCA}. 
We can see that using the same input between keys and values in MH-CA (MH-CA $^{*}$) requires more parameters and FLOPs but cannot bring further performance gains. 
It illustrates the effectiveness of our efficient strategy in MCA.

\begin{figure}[htb]
   \centering
   \begin{subfigure}[htb]{0.497\textwidth}
      \includegraphics[width=\textwidth]{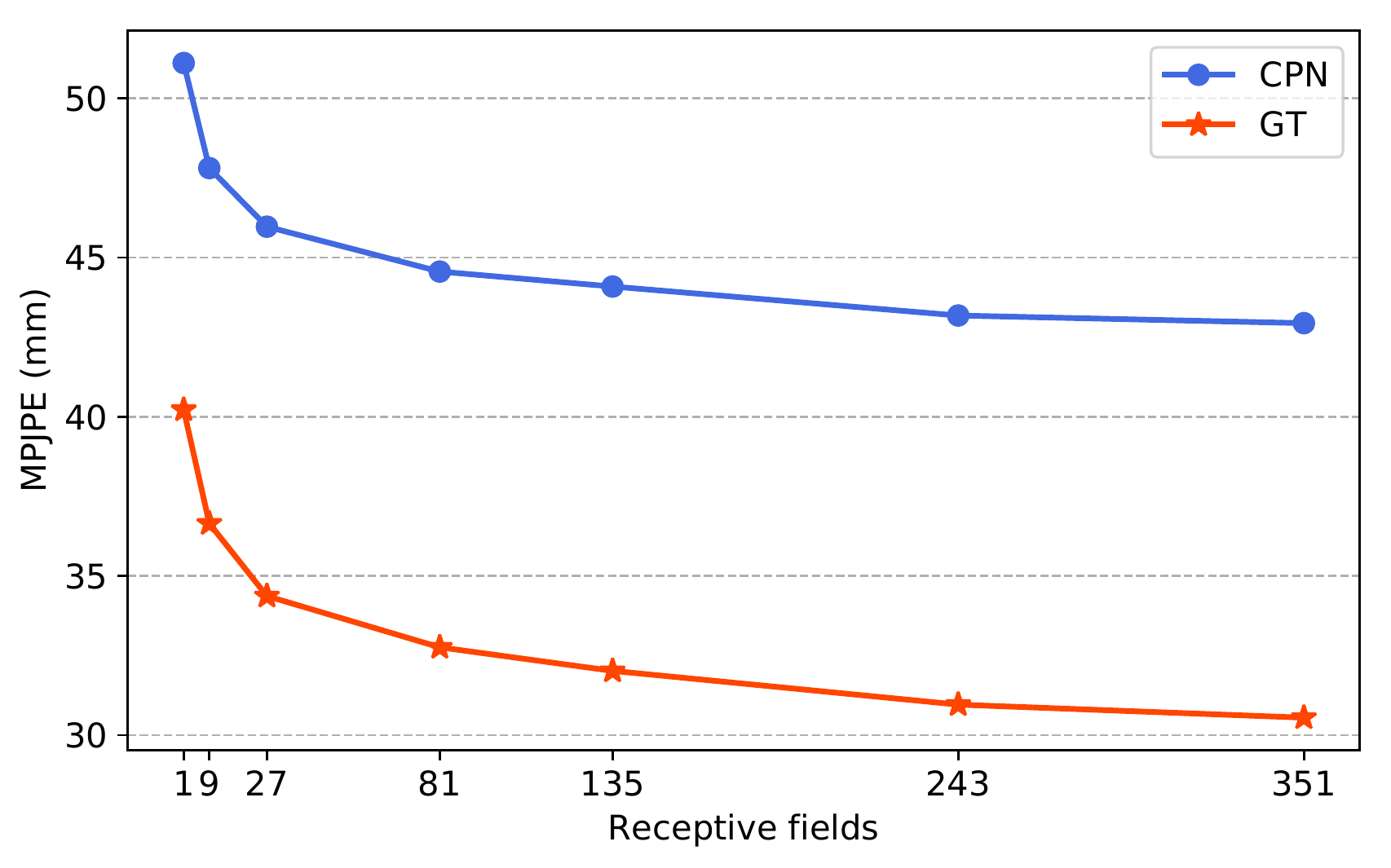}
      \caption{Different receptive fields under MPJPE.}
      \label{fig:frames}
   \end{subfigure}
   \begin{subfigure}[htb]{0.497\textwidth}
      \includegraphics[width=\textwidth]{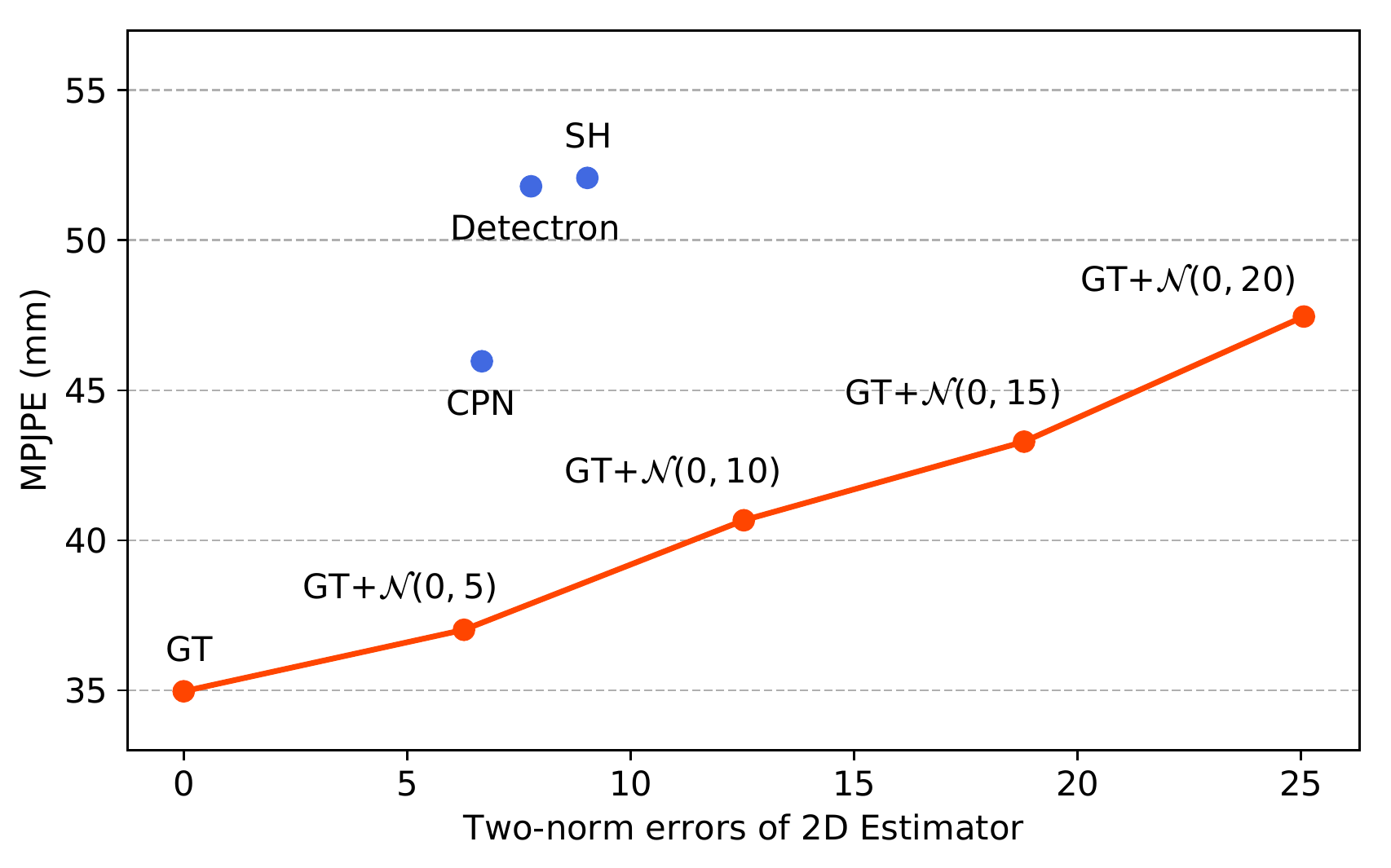}
      \caption{Different 2D detections under MPJPE.}
      \label{fig:2D_detections}
   \end{subfigure}
   \caption
   {
   \textbf{(a)} Ablation studies on different receptive fields of our method on Human3.6M under MPJPE metric. 
   \textbf{(b)} The effect of 2D detections on Human3.6M under MPJPE. 
   Here, $\mathcal{N}(0, \sigma)$ represents the Gaussian noise with mean zero and $\sigma$ is the standard deviation. 
   (CPN) - Cascaded Pyramid Network; (SH) Stack Hourglass; (GT) - 2D ground truth.  
   }
   \label{fig:frames and 2D}
   \VspaceL
\end{figure}

\noindent
\textbf{Impact of Receptive Fields.}
For the video-based 3D human pose estimation task, 
the number of receptive fields directly influences the estimation results. 
Figure~\ref{fig:frames and 2D} (a) shows the results of our model with different receptive fields (between 1 and 351) on Human3.6M. 
Increasing the receptive field can improve the result under both CPN and GT 2D pose inputs, which demonstrates the great power of our method in long-range dependency modeling with a long input sequence. 

\noindent
\textbf{Impact of 2D Detections.}
To show the effectiveness of our method on different 2D pose detectors, we carry out experiments with the detections from Stack Hourglass (SH)~\cite{newell2016stacked}, Detectron~\cite{pavllo20193d}, and CPN~\cite{chen2018cascaded}. 
In addition, to evaluate the robustness of our method to various levels of noise, we also conduct experiments on 2D ground truth plus different levels of additive Gaussian noise. 
The results are shown in Figure~\ref{fig:frames and 2D} (b). 
It can be observed that the curve has a nearly linear relationship between MPJPE of 3D poses and two-norm errors of 2D poses. 
These experiments validate both the effectiveness and robustness of our proposed method. 

\section{Additional Visualization Results}
\begin{figure*}[htb]
   \centering
   \includegraphics[width=0.952 \linewidth]{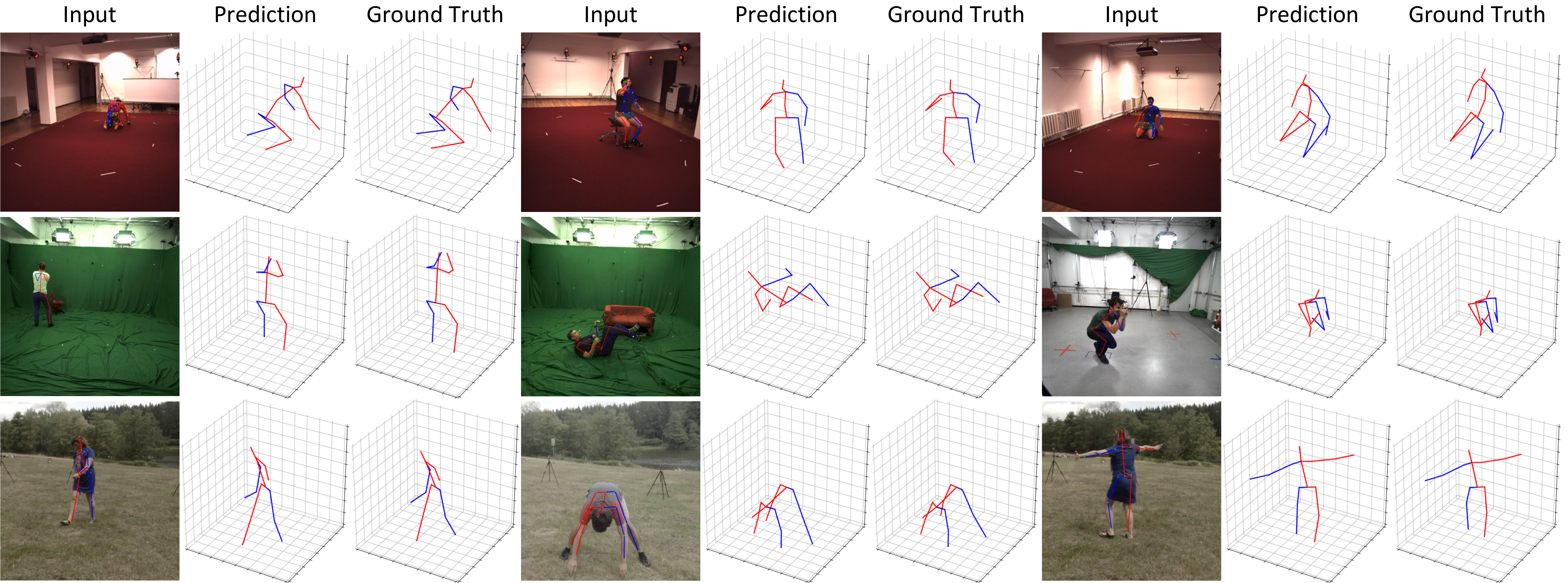}
   \caption
   {Qualitative results of our proposed method on Human3.6M dataset (first 1 row) and MPI-INF-3DHP dataset (last 2 rows). 
   }
   \label{fig:dataset}
   \vspace{-0.1cm}
\end{figure*}

\begin{figure*}[htb]
   \centering
   \includegraphics[width=0.952\linewidth]{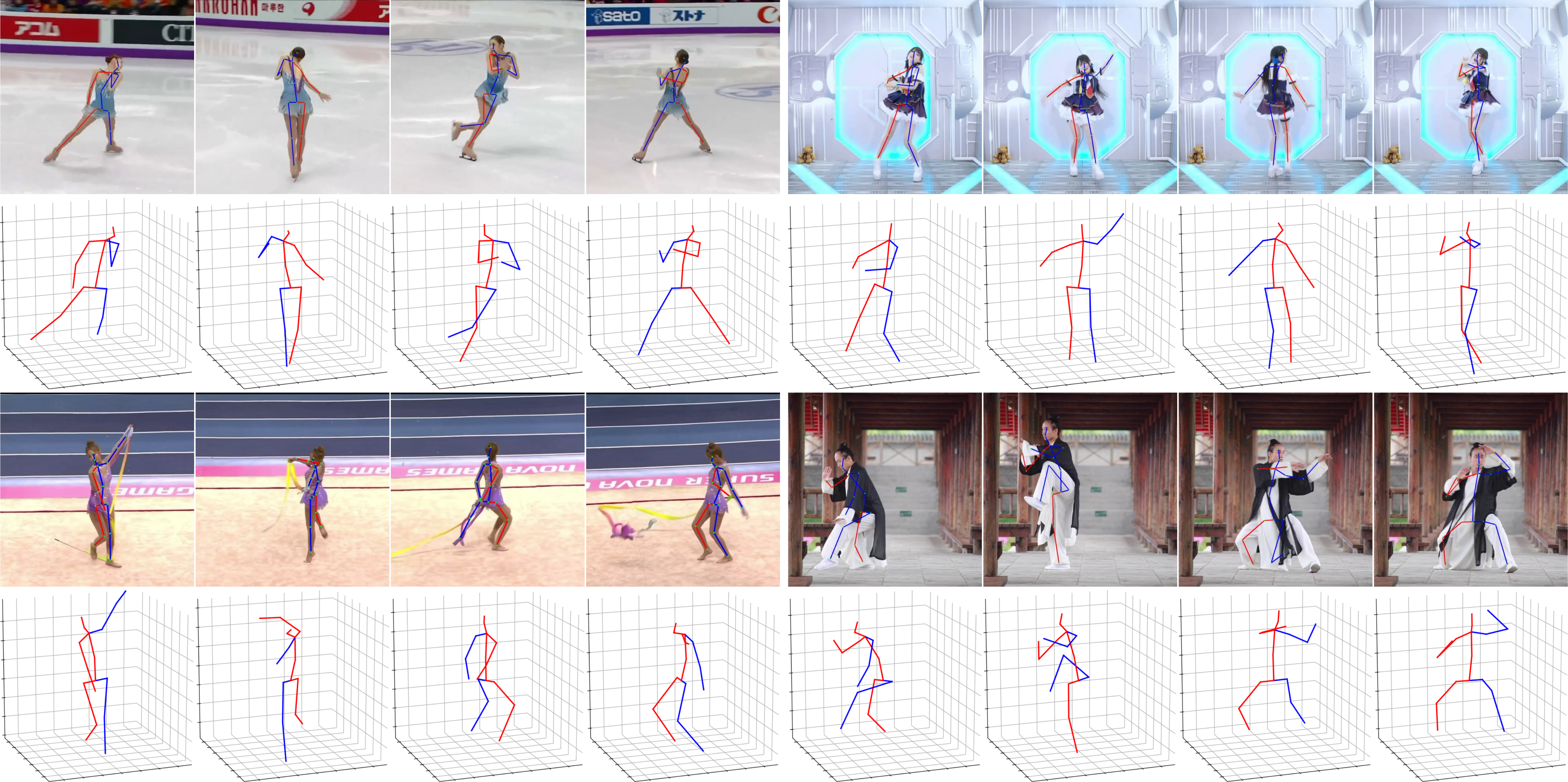}
   \caption
   {Qualitative results of our proposed method on challenging in-the-wild videos. 
   }
   \label{fig:wild}
   \vspace{-0.1cm}
\end{figure*}

\begin{figure*}[htb]
   \centering
   \includegraphics[width=0.952\linewidth]{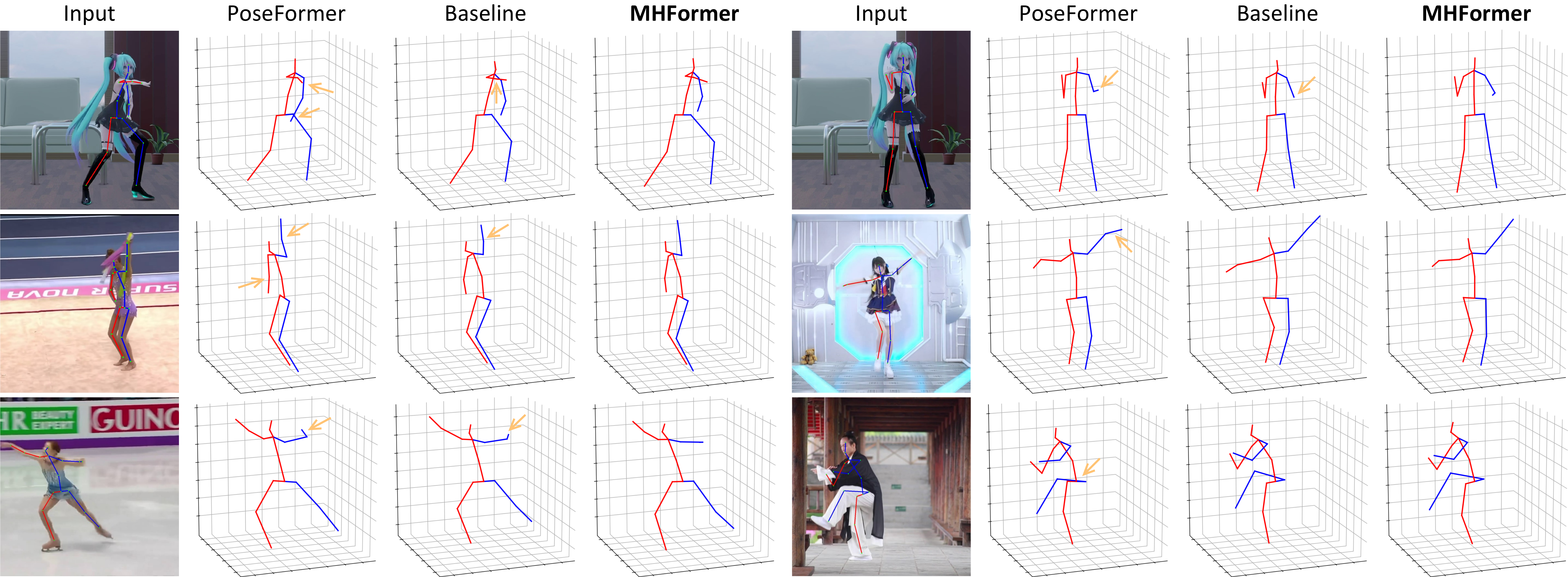}
   \caption
   {Qualitative comparison among the proposed method (MHFormer), the baseline method, and the previous state-of-the-art method (PoseFormer)~\cite{poseformer} on challenging wild videos. 
   Wrong estimations are highlighted by yellow arrows. 
   }
   \label{fig:wild_compare}
   \vspace{-0.4cm}
\end{figure*}

\noindent \textbf{3D Reconstruction Visualization.}
Figure~\ref{fig:dataset} and Figure~\ref{fig:wild} show qualitative results of our method on Human3.6M dataset, MPI-INF-3DHP dataset, and challenging in-the-wild videos. 
Moreover, Figure~\ref{fig:wild_compare} shows the qualitative comparison with the baseline method and the previous state-of-the-art method (PoseFormer~\cite{poseformer}) on some wild videos. 
It can be seen that our method can produce more accurate and reasonable 3D poses, especially when the human action is complex and rare. 

\begin{figure*}[htb]
   \centering
   \includegraphics[width=1.\linewidth]{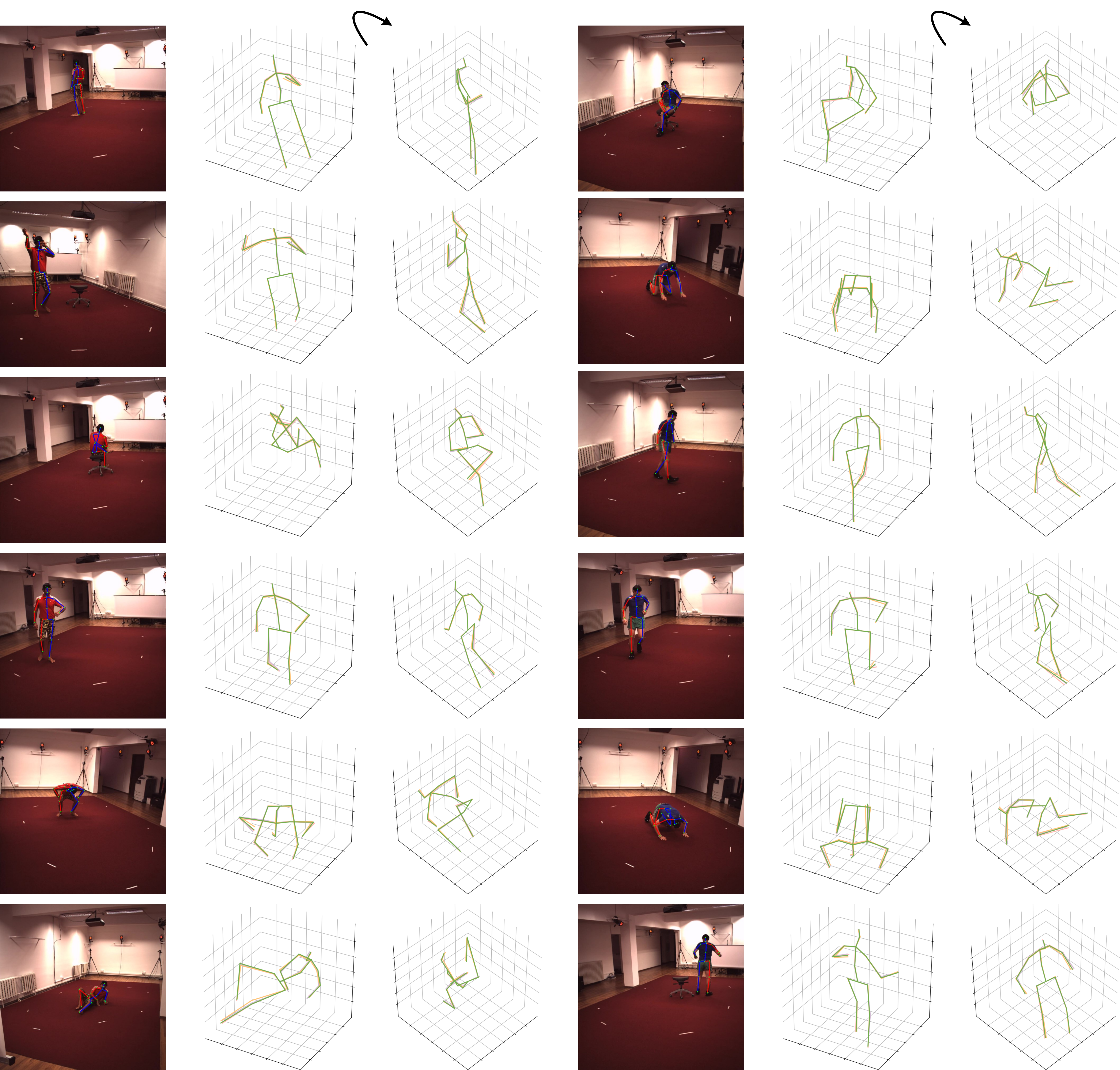}
   \caption
   {Diverse 3D pose hypotheses generated by MHFormer. 
   For easy illustration, we color-code the hypotheses to show the difference among them, and the hypotheses are shown from two perspectives. 
   Green colored 3D pose corresponds to the final synthesized estimation of our method. 
   }
   \label{fig:hypotheses_all}
\end{figure*}

\noindent \textbf{Hypothesis Visualization.}
For visualization purposes, we add additional regression layers and finetune our model to output intermediate hypotheses. 
Figure~\ref{fig:hypotheses_all} shows the visualization results of intermediate 3D pose hypotheses generated by our proposed method. 
We can see that our MHFormer can generate different plausible 3D pose solutions, especially for ambiguous body parts with depth ambiguity, self-occlusion, and 2D detector uncertainty. 

\begin{figure*}[!htb]
   \centering
   \begin{subfigure}[htb]{0.33\textwidth}
      \includegraphics[width=\textwidth]{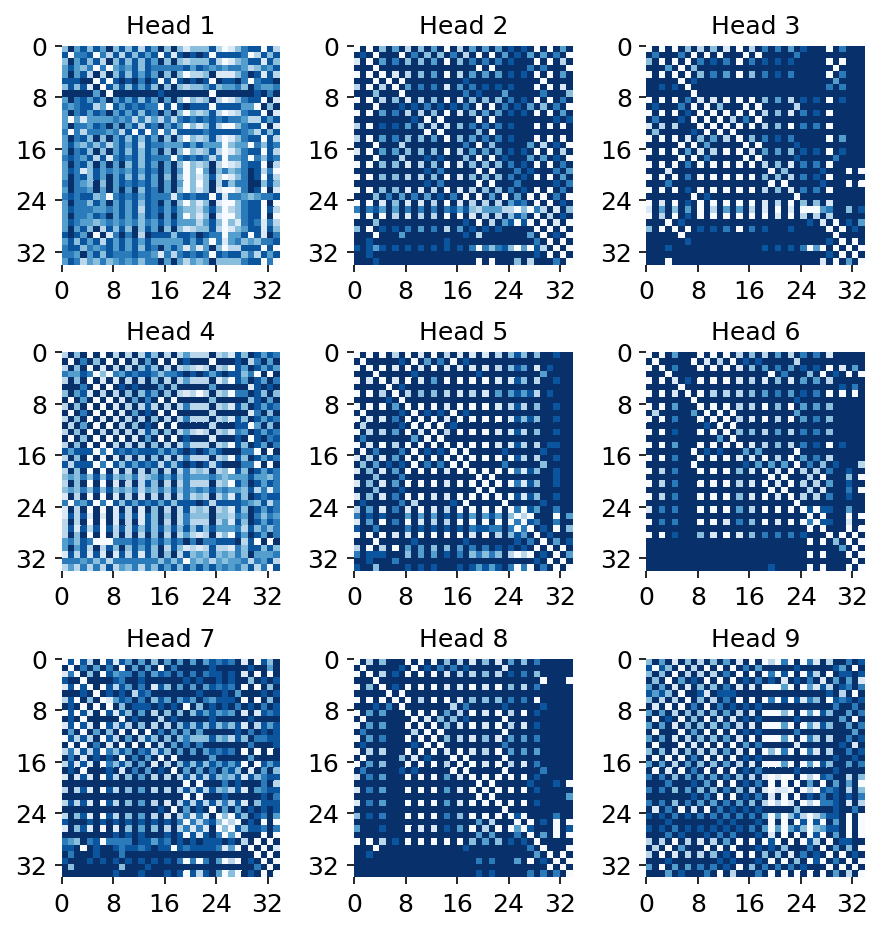}
      \caption{Hypothesis 1}
      \label{fig:spatial_1}
   \end{subfigure}
   \begin{subfigure}[htb]{0.33\textwidth}
      \includegraphics[width=\textwidth]{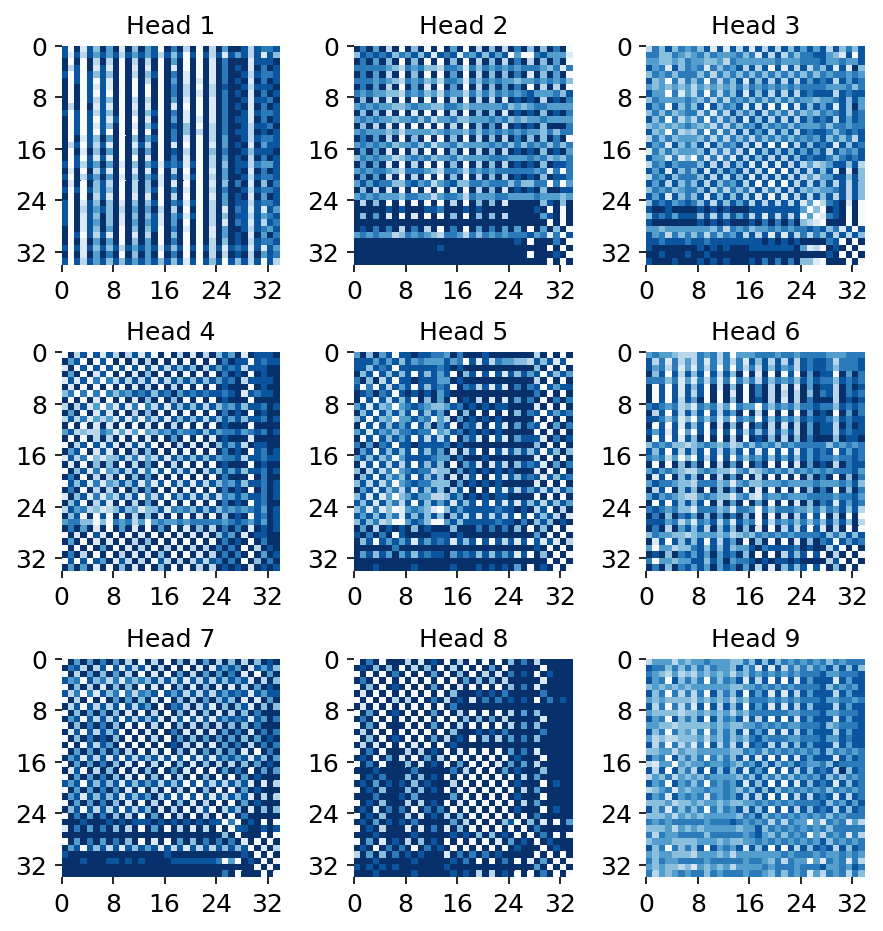}
      \caption{Hypothesis 2}
      \label{fig:spatial_2}
   \end{subfigure}
   \begin{subfigure}[htb]{0.33\textwidth}
      \includegraphics[width=\textwidth]{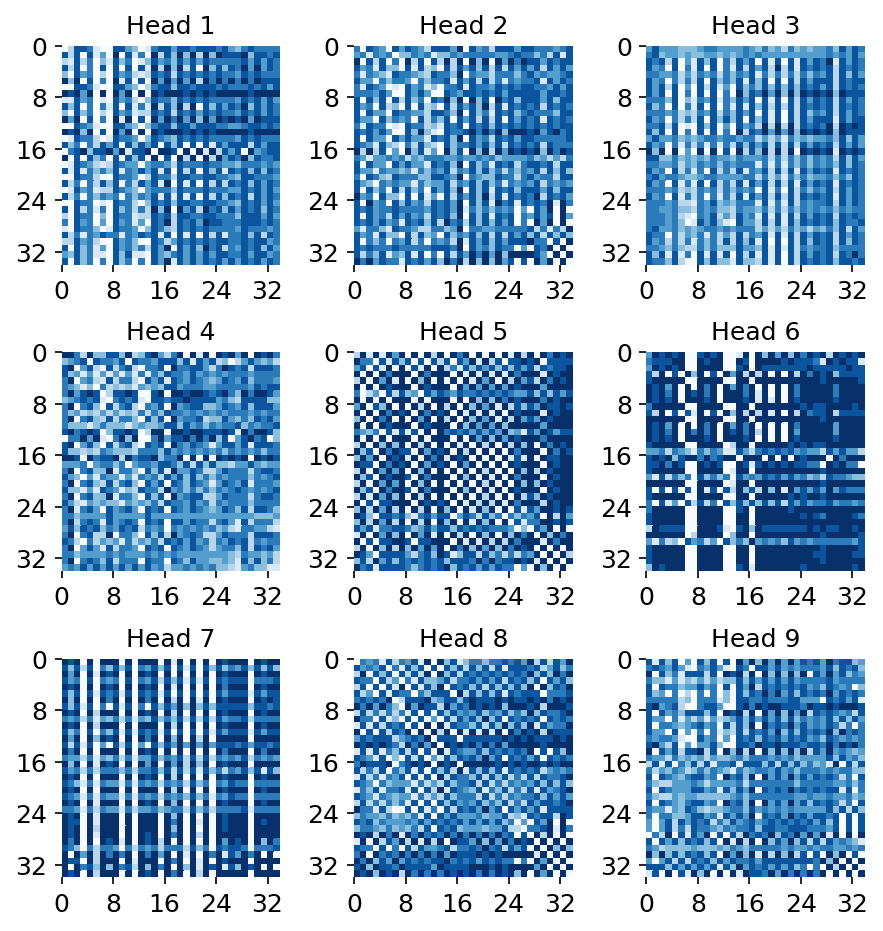}
      \caption{Hypothesis 3}
      \label{fig:spatial_3}
   \end{subfigure}
   \caption
   {
      Multi-head attention maps (9 heads) from the Multi-Hypothesis Generation (MHG) module of our 351-frame model with 3 different hypotheses. 
      The brighter color indicates a stronger attention value. 
   }
   \label{fig:spatial}
\end{figure*}

\begin{figure*}[!htb]
   \centering
   \begin{subfigure}[htb]{0.497\textwidth}
      \includegraphics[width=\textwidth]{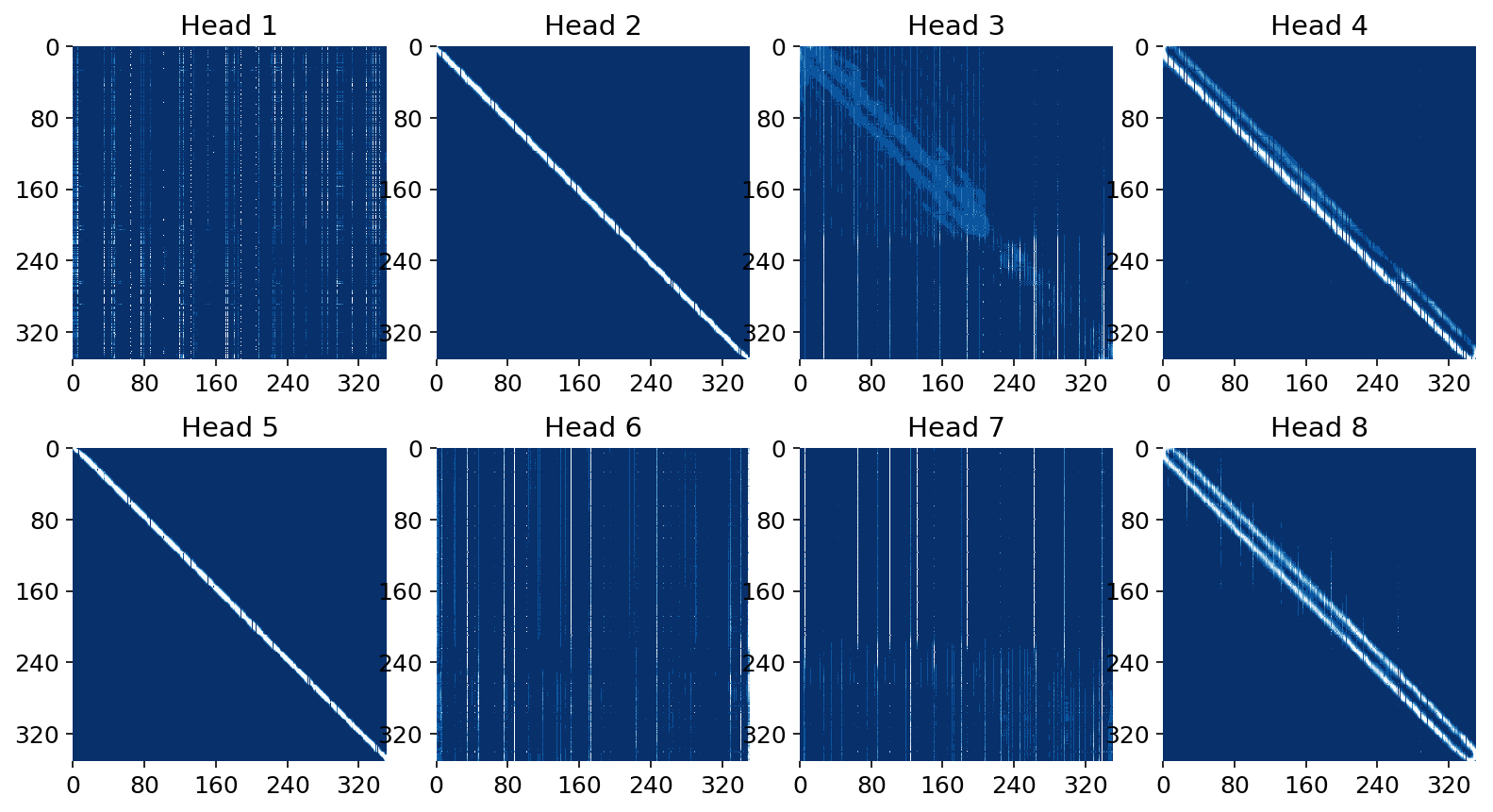}
      \caption{Hypothesis 1}
      \label{fig:temporal_1}
   \end{subfigure}
   \begin{subfigure}[htb]{0.497\textwidth}
      \includegraphics[width=\textwidth]{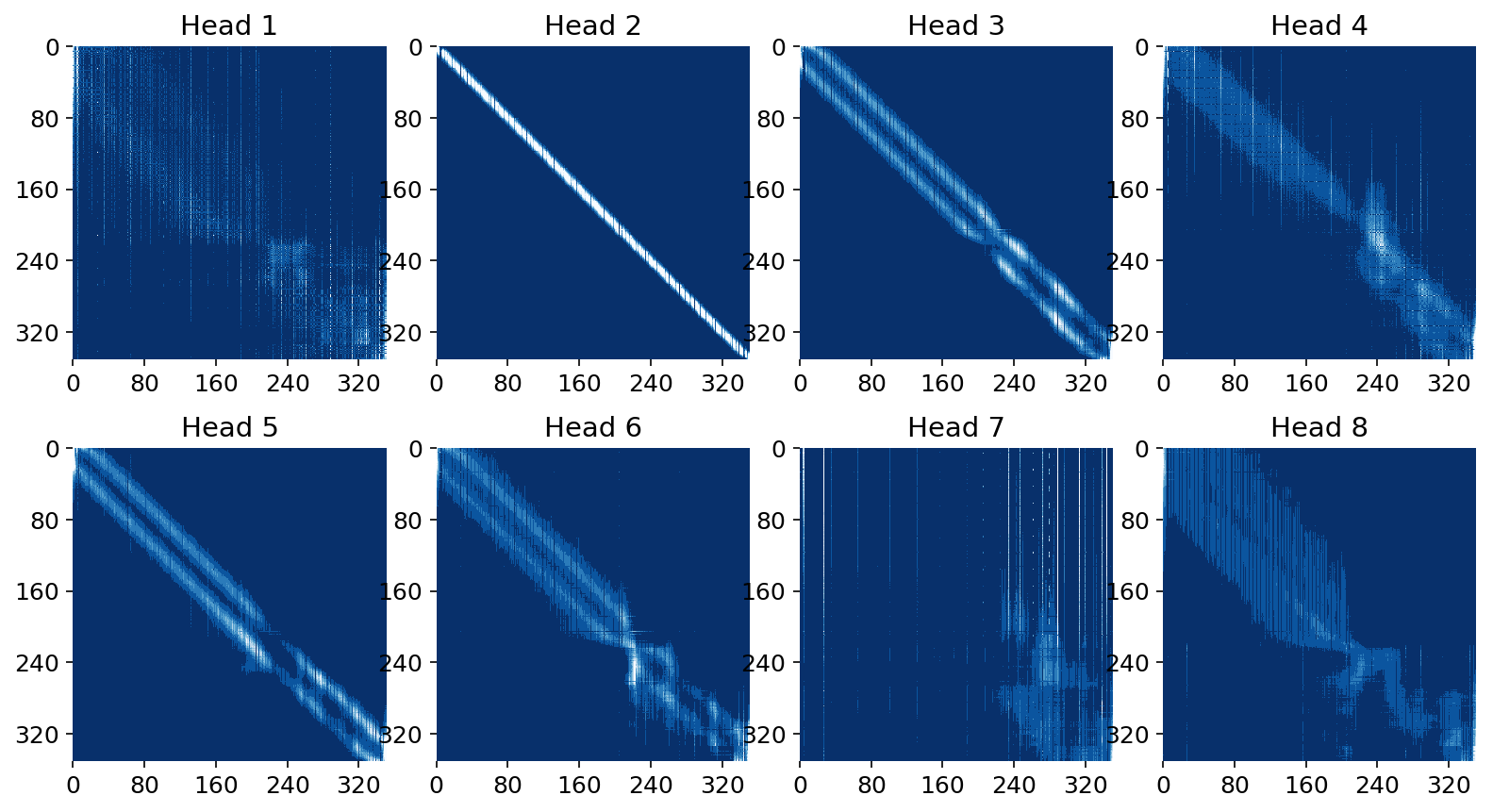}
      \caption{Hypothesis 2}
      \label{fig:temporal_2}
   \end{subfigure}
   \begin{subfigure}[htb]{0.497\textwidth}
      \includegraphics[width=\textwidth]{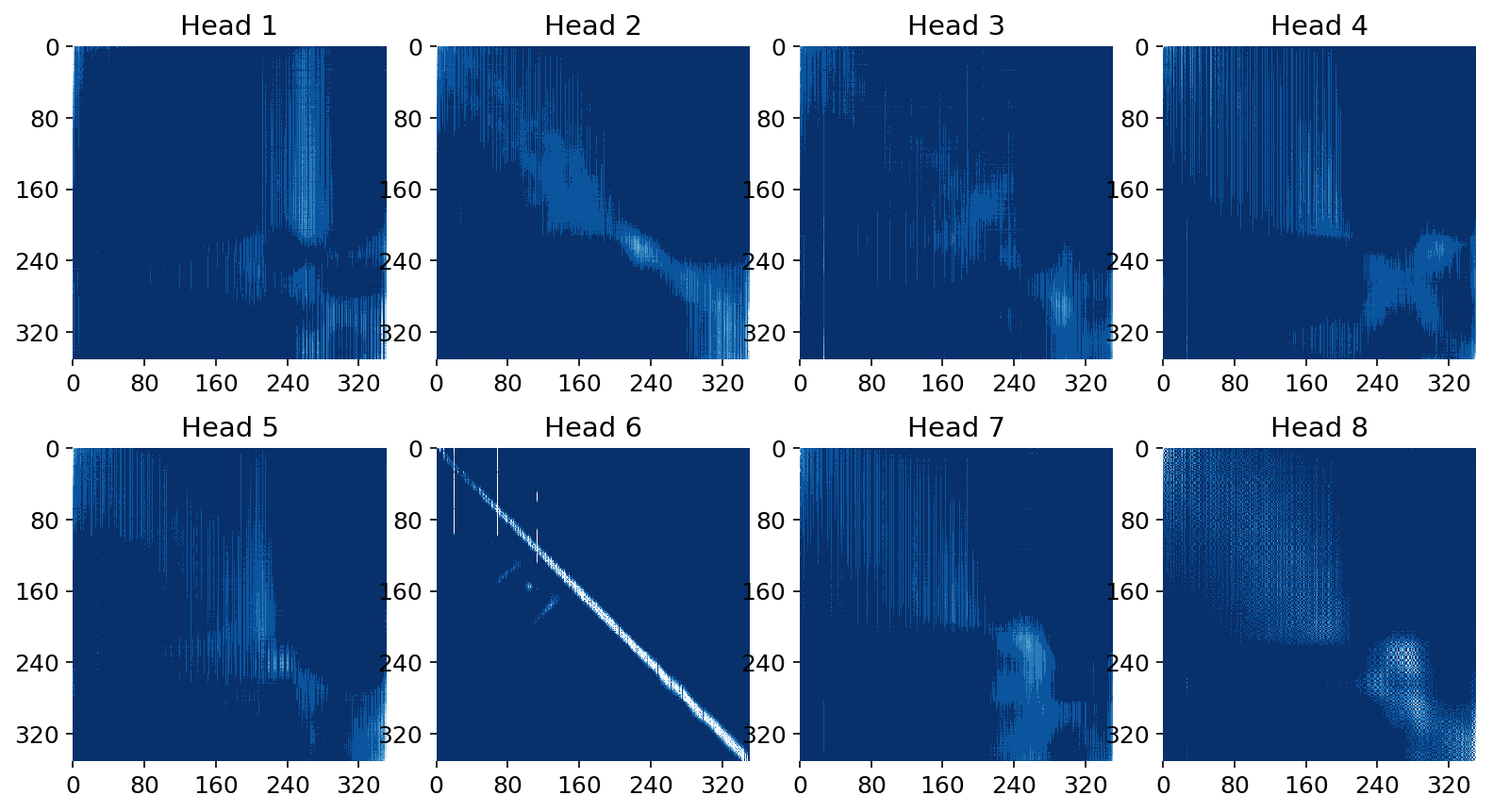}
      \caption{Hypothesis 3}
      \label{fig:temporal_3}
   \end{subfigure}
   \caption
   {
      Multi-head attention maps (8 heads) from the Self-Hypothesis Refinement (SHR) module of our 351-frame model with 3 different hypotheses. 
      The brighter color indicates a stronger attention value. 
   }
   \label{fig:temporal}
\end{figure*}

\noindent \textbf{Attention Visualization.}
Visualization results of the multi-head attention maps of the first layers from the Multi-Hypothesis Generation (MHG) module and Self-Hypothesis Refinement (SHR) module (351-frame model with 3 hypotheses) are shown in Figure~\ref{fig:spatial} and Figure~\ref{fig:temporal}, respectively. 
It can be found that the maps of multiple hypotheses contain diverse patterns and semantics. 
This indicates multiple representations in our method actually learn various modal information of pose hypotheses. 

\end{document}